\documentclass{ecai}

\usepackage{times}
\usepackage{latexsym}

\usepackage[utf8]{inputenc}
\usepackage{amsmath,amsfonts,amssymb,amsthm}
\usepackage{graphicx}
\usepackage{stmaryrd}
\usepackage{microtype}
\usepackage{url}
\usepackage{enumitem}
\usepackage{IEEEtrantools}
\usepackage{xcolor}
\binoppenalty=\maxdimen
\relpenalty=\maxdimen

\begin{document}
\title{An ASP semantics for Constraints involving\\ Conditional Aggregates}

\author{%
  Pedro Cabalar\institute{University of Corunna, Spain}
  \and
  Jorge Fandinno\institute{University of Potsdam, Germany}
  \and
  Torsten Schaub$^2$
  \and Philipp Wanko$^2$ }

\newtheorem{definition}{Definition}
\newtheorem{proposition}{Proposition}
\newtheorem{corollary}{Corollary}
\newtheorem{observation}{Observation}
\newtheorem{lemma}{Lemma}
\newtheorem{example}{Example}
\newtheorem{thm}{Theorem}

\newcommand{\den}[1]{\llbracket \, #1 \, \rrbracket}
\newcommand{\ctermm}[3]{\ensuremath{{#1|#2}{:\,}#3}}
\newcommand{\cterm}[3]{\ensuremath{(\ctermm{#1}{#2}{#3})}} 
\newcommand{\close}[1]{\ensuremath{#1\raisebox{1pt}{$\uparrow$}}}

\newcommand{\htag}[2]{\ensuremath{\mathit{#1}\agg{#2}}}
\newcommand{\htagg}[4]{\ensuremath{\mathit{#1}\agg{#2}\mathrel{#3}#4}}
\newcommand{\htaggg}[4]{\ensuremath{\mathit{#1} {#2 }\mathrel{#3}#4}}
\newcommand{\vals}[2]{\ensuremath{\mathcal{V}_{#1,#2}}}
\newcommand{\eval}[2]{\ensuremath{\mathit{eval}}_{\langle #1,#2\rangle}}
\newcommand{\evalgz}[2]{\ensuremath{\mathit{eval}^{\mathtt{gz}}_{\langle #1,#2\rangle}}}
\newcommand{\evalf}[2]{\ensuremath{\mathit{eval}^{\mathtt{f}}_{\langle #1,#2\rangle}}}
\newcommand{\evalcl}[1]{\ensuremath{\mathit{eval}_{#1}}}

\newcommand{\evall}[1]{\ensuremath{\mathit{eval}}_{#1}}
\newcommand{\evallgz}[1]{\ensuremath{\mathit{eval}^{\mathtt{gz}}_{#1}}}

\newcommand{\modelscl}{\ensuremath{\models_{cl}}}
\newcommand{\eqdef}{\ensuremath{\mathbin{\raisebox{-1pt}[-3pt][0pt]{$\stackrel{\mathit{def}}{=}$}}}} 

\newcommand{\sysfont}{\textit}
\newcommand{\clingo}{\sysfont{clingo}}
\newcommand{\clingoDL}{\clingo{\small\textnormal{[}\textsc{DL}\textnormal{]}}}
\newcommand{\dlprogram}{\textit{DL}-program}
\newcommand{\code}[1]{\texttt{#1}}
\newcommand{\ground}{variable-free}
\newcommand{\modelsgz}{\models_{\mathtt{gz}}}
\newcommand{\modelsf}{\models_{\mathtt{f}}}
\newcommand{\modelss}{\models_{\mathtt{s}}}
\newcommand{\Atoms}{\mathit{Atoms}}
\newcommand{\Head}{\mathit{H}}
\newcommand{\HeadA}{\mathit{\Head_A}}
\newcommand{\HeadB}{\mathit{\Head_B}}
\newcommand{\Headp}{\mathit{\Head^+}}
\newcommand{\Headn}{\mathit{\Head^-}}
\newcommand{\Body}{\mathit{B}}
\newcommand{\var}{\mathit{var}}
\newcommand{\Cond}{\mathit{Cond}}
\newcommand{\tuple}[1]{\langle #1 \rangle}
\newcommand{\restr}[2]{{#1|}_{\hspace{-1pt}#2}}
\newcommand{\df}[1]{\mathit{def}(#1)}
\newcommand{\HTC}{\ensuremath{\mathit{HT}_{\!C}}} 
\newcommand{\HT}{\ensuremath{\mathit{HT}}}
\newcommand{\X}{\ensuremath{\mathcal{X}}}
\newcommand{\D}{\ensuremath{\mathcal{D}}}
\newcommand{\Du}{\ensuremath{\mathcal{D}_{\mathbf{u}}}}
\newcommand{\C}{\ensuremath{\mathcal{C}}}
\newcommand{\F}{\ensuremath{\mathcal{F}}}
\newcommand{\I}{\ensuremath{\mathcal{I}}}
\newcommand{\A}{\ensuremath{\mathcal{A}}}
\newcommand{\Piref}[1]{\Pi_{\ref{#1}}}
\newcommand{\vars}[1]{\ensuremath{\mathit{vars}(#1)}}
\newcommand{\At}[1]{\ensuremath{\mathit{At}(#1)}}
\newcommand{\val}{\ensuremath{v}} 
\newcommand{\f}{\ensuremath{\mathbf{f}}} 
\newcommand{\undefined}{\ensuremath{\mathbf{u}}} 
\newcommand{\ass}[3]{#1 := #2 \, .. \, #3}
\newcommand{\LC}{\ensuremath{\mathit{LC}}}
\newcommand{\DF}{\ensuremath{\boldsymbol{D\hspace{-1.2pt}F}}} 
\newcommand{\agg}[1]{\ensuremath{\dot{\{}#1\dot{\}}}} 
\newcommand{\LX}{\ensuremath{\mathbb{X}}}
\newcommand{\HU}{\ensuremath{\mathbb{U}}}
\newcommand{\Gra}{\ensuremath{\mathit{G}^a}}
\newcommand{\Gr}{\ensuremath{\mathit{G}}}
\newcommand{\Def}{\delta}
\newcommand{\grsep}{\,\big|\hspace{-3pt}\big|\hspace{-3pt}\big|\,}

\def\sumf{\mathit{sum}}


\maketitle
\bibliographystyle{ecai} 

\begin{abstract}
  We elaborate upon the formal foundations of hybrid Answer Set Programming (ASP)
  and extend its underlying logical framework with aggregate functions over constraint values and variables.
  This is achieved by introducing the construct of conditional expressions,
  which allow for considering two alternatives while evaluating constraints.
  Which alternative is considered is interpretation-dependent and
  chosen according to an associated condition.
  We put some emphasis on logic programs with linear constraints
  and show how common ASP aggregates can be regarded as particular cases of so-called conditional linear constraints.
  Finally,
  we introduce a polynomial-size, modular and faithful translation
  from our framework into regular (condition-free) Constraint ASP,
  outlining an implementation of conditional aggregates on top of existing hybrid
  ASP solvers.
\end{abstract}
%

%
\section{Introduction}\label{sec:introduction}

Many real-world applications have a heterogeneous nature.
Let it be in
bio-informatics~\cite{frscscsiwa18a},
hardware synthesis~\cite{newascha18b},
or train scheduling~\cite{abjoossctowa19a},
all cited ones consist of genuine qualitative and quantitative constraints.
While the former often account for topological requirements, like reachability,
the latter usually address (fine-grained) time or resource requirements.

The hybrid nature of such applications has led to mixed solving technology,
foremost in the area of Satisfiability modulo Theories (SMT;~\cite{niolti06a}).
Meanwhile,
the need for hybridization has also prompted similar approaches in neighboring areas such as Answer Set Programming (ASP;~\cite{lifschitz08b}).
However,
while ASP solving technology is at eye-height with that of SAT and SMT,
its true appeal lies in its high-level modeling language building on a non-monotonic semantics.
Among others, this allows for expressing defaults and an easy formulation of reachability.
When it comes to extending ASP with foreign reasoning methods,
the design often follows the algorithmic framework of SMT and leaves semantic aspects behind.
For instance, a popular approach is to combine ASP with Constraint Processing (CP;~\cite{dechter03a}),
also referred to as Constraint ASP (CASP;~\cite{lierler14a}).
This blends non-monotonic aspects of ASP with monotonic ones of CP but
fails to provide a homogeneous representational framework.
In particular, the knowledge representation capabilities of ASP, like defaults and aggregates, remain inapplicable to constraint variables.

We addressed this in~\cite{cakaossc16a} by integrating ASP and CP in the uniform semantic framework called \emph{Here-and-There with constraints}~(\HTC).
The idea is to rebuild the logic of ASP from constraint atoms encapsulating arbitrary foreign constraints.
This relies upon the logic of Here-and-There (\HT;~\cite{heyting30a}) along with
its non-monotonic extension, called Equilibrium Logic~\cite{pearce96a}.
Although \HTC\ offers a uniform representation, for instance, featuring defaults for constraint variables,
it still lacks an essential element of ASP's modeling language, namely, aggregates with conditional elements.
This issue is addressed in the paper at hand.
As an example,
consider the hybrid ASP rule%
\footnote{We put dots on top of braces, viz.~``$\agg{ \dotsc }$'', to indicate \emph{multisets}.}
\begin{gather}\label{eq:tax.sum}
  \mathit{total}(R) := \mathit{sum}\agg{ \, \mathit{tax}(P) : \mathit{lives}(P,R) \,  } \ \leftarrow \ \mathit{region}(R)
  \quad
\end{gather}
gathering the total tax revenue of each region $R$ by summing up the tax liabilities of the region's residents, $P$.
As a matter of fact,
the calculation of tax liability is highly complex, and relies on defaults and discounts to address incomplete information,
which nicely underlines the need for non-monotonic constraint variables.
Once instantiated,
$\mathit{lives}(P,R)$ and $\mathit{region}(R)$ are propositional atoms,
while the entire rule head is regarded as a constraint atom,
whose actual meaning eludes the ASP system
(just as in lazy SMT~\cite{niolti06a}).
The aggregate function $\mathit{sum}$ is applied to a set of conditional expressions of the form
\mbox{$\mathit{tax}(P) : \mathit{lives}(P,R)$}.
This makes sure that each instantiated rule gathers only taxes accrued by the inhabitants of the respective region~$R$.
Although such an aggregation is very appealing from a modeling perspective,
it leaves us with several technical challenges.
First, the set of arguments is context-dependent and thus a priori unknown.
Second, the identification of valid arguments necessitates the evaluation of Boolean conditions within foreign and thus opaque constraint atoms.

Accordingly,
we start by developing a formal account of \emph{conditional expressions};
they allow us to consider two alternatives while evaluating constraints.
Which alternative is considered is interpretation-dependent and
chosen according to the evaluation of a given condition,
expressed as a logical formula.
As a case-study,
we focus on a syntactic fragment extending logic programs with linear constraints.
We show that $\mathit{sum}$, $\mathit{count}$, $\mathit{min}$ and $\mathit{max}$ aggregate atoms in ASP constitute a special case of conditional linear constraints.
Interestingly,
our framework refrains from imposing the treatment of aggregate atoms as a whole;
rather considering them as sub-expressions that can be combined with other arithmetic operations,
leaving aggregate atoms a simple particular case.
Finally, we develop a translation of programs with conditional expressions into CASP,
which is itself condition-free.
This enables the use of off-the-shelf CASP solvers as back-ends for implementing our approach.
In this way, our translation allows us to delegate the responsibilities for evaluating constraints with conditional expressions:
an ASP solver is in charge of evaluating conditions, while an associated CP solver only deals with condition-free constraints.
%

%
\section{The Logic of Here-and-There with Constraints}
\label{sec:ht:c}

The syntax of \HTC\ is based on a set of (constraint) variables $\mathcal{X}$
and constants or domain values%
\footnote{%
  Formally, we assume unique names for constants and use the same symbol for the constant name and the domain element.
  Note that, from a purely logical point of view, constraint variables are
  first order variables or $0$-ary functions for which the standard name assumption is not assumed.
  As a result, its associated value is interpretation dependent.}
from some non-empty set $\mathcal{D}$.
A \emph{constraint atom} is some expression that is used to relate values of variables and constants
according to the atom's semantics.
We use $\mathcal{C}$ to denote the set of all constraint atoms.

Most useful constraint atoms have a syntax defined by some grammar or regular pattern:
for instance, difference constraints are expressions of the form ``$x - y \leq d$'',
where $x$ and $y$ are variables from~$\mathcal{X}$ and $d$ is a constant from $\mathcal{D}$.
At the most general level, however, we only require that the constraint atom is represented by some string of symbols
(we generally call \emph{expression}),
possibly, of infinite length.
Apart from operators or punctuation symbols,
this string may name some
variables from~$\mathcal{X}$,
constants from~$\mathcal{D}$, and, for convenience, also
a special symbol $\undefined \notin \mathcal{D}$ that stands for \emph{undefined}.
We define the extended domain $\mathcal{D}_{\undefined} \eqdef \mathcal{D} \cup \{\undefined\}$.
The set $\mathit{vars}(c) \subseteq \mathcal{X}$
collects all variables occurring in constraint $c$.
We sometimes refer to a constraint atom using the notation $c[s]$ meaning that
the expression for $c$ contains some distinguished occurrence of subexpression $s$.
We further write $c[s/s']$ to represent the syntactic replacement in $c$ of subexpression~$s$ by~$s'$ as usual.
We assume that~$c[x/d] \in \mathcal{C}$ for every constraint atom~$c[x] \in \C$, variable~$x \in \X$ and~$d \in \Du$.
That is, replacing a variable by any element of the extended domain results in a syntactic valid constraint atom.

A \emph{valuation} $v$ over $\mathcal{X},\mathcal{D}$ is some total function
\mbox{$v:\mathcal{X}\rightarrow\mathcal{D}_{\undefined}$}
where
\mbox{$v(x)=\undefined$}
represents that variable $x$ is left undefined.
Moreover, if $X \subseteq \mathcal{X}$ is a subset of variables, valuation $v|_X: X\rightarrow\mathcal{D}_{\undefined} $ stands for the projection of $v$ on $X$.
A valuation $v$ can be alternatively represented as the set
\mbox{$\{ (x,v(x)) \mid x \in \mathcal{X}, v(x)\in\mathcal{D}\}$},
excluding pairs of form $(x,\undefined)$ from the set.
This representation allows us to use standard set inclusion for comparison.
We thus write $v\subseteq v'$ to mean that
\(
\{ (x,v (x)) \mid x \in \mathcal{X}, v (x)\in\mathcal{D}\}
\subseteq
\{ (x,v'(x)) \mid x \in \mathcal{X}, v'(x)\in\mathcal{D}\}
\).
This is equivalent to: $v(x) \in \mathcal{D}$ implies $v'(x)=v(x)$ for all $x \in \X$.
We also allow for applying valuations $v$ to fixed values,
and so extend their type to
\(
v:\mathcal{X}\cup \mathcal{D}_{\undefined}\rightarrow\mathcal{D}_{\undefined}
\)
by fixing $v(d) = d$ for any $d \in \mathcal{D}_\undefined$.
The set of all valuations over $\mathcal{X},\mathcal{D}$ is denoted by $\mathcal{V}_{\mathcal{X},\mathcal{D}}$
and $\mathcal{X},\mathcal{D}$ dropped whenever clear from context.

We define the semantics of constraint atoms via \emph{denotations},
which are functions
\(
\den{\cdot}:\mathcal{C}\rightarrow 2^{\mathcal{V}}
\),
mapping each constraint atom to a set of valuations.
We require denotations~$\den{\cdot}$ to satisfy the following properties
for all
$c\in\mathcal{C}$,
$x\in\mathcal{X}$, and
any~$v,v' \in \mathcal{V}$:
\begin{enumerate}
\item   $v \in \den{c}$ and $v \subseteq v'$ imply $v' \in \den{c}$,
  \label{den:prt:0}
\item $v \in \den{c}$ implies $v \in \den{c[x/v(x)]}$,
  \label{den:prt:1}
\item if $v(x)=v'(x)$ for all $x \in \mathit{vars}(c)$ then $v \in \den{c}$ iff $v' \in \den{c}$.
  \label{den:prt:2}
\end{enumerate}
Intuitively,
Condition~\ref{den:prt:0} makes constraint atoms behave monotonically.
Condition~\ref{den:prt:1} stipulates that denotations respect the role of variables as placeholders for values, that is,
replacing variables by their assigned value does not change how an expression is evaluated.
Condition~\ref{den:prt:2} asserts that the denotation of $c$ is fixed by combinations of values for $\vars{c}$,
while all other variables are irrelevant and may freely vary.

The flexibility of syntax and semantics of constraint atoms allows us to capture entities across different theories.
For instance, assuming a value $\mathbf{t} \in \D$ for representing the truth value \emph{true},
Boolean propositions can be modelled via constraint atoms having a denotation
\(
\den{a}=\{v\in\mathcal{V}\mid v(a)=\mathbf{t}\}.
\)
As we mention above, we can cover different variants of linear equations.
For example,
difference constraint of the form ``$x-y\leq d$''
can be captured via constraint atoms of the same syntax ``$x-y\leq d$'' whose denotation is the expected\
\begin{align*}
  \den{\text{``}x-y\leq d\text{''}}
  =
  \{v\in\mathcal{V}\mid v(x),v(y), d\in \mathbb{Z}, v(x)-v(y)\leq d\}
  \ ,
\end{align*}
where $\vars{\text{``}x-y\leq d\text{''}} = \{x,y\} \subseteq \mathcal{X}$ and $d\in\D$.
Note that this difference constraint can only be satisfied when $x$ and $y$ hold an integer value and $d\in \mathbb{Z}$.
For clarity, we simply remove quotes, when clear from the context.
In what follows,
we assume that integers and the truth value~$\mathbf{t}$ are part of the domain, that is, $\{ \mathbf{t} \} \cup \mathbb{Z} \subseteq \mathcal{D}$.

A formula $\varphi$ over $\mathcal{C}$ is defined as
\begin{align*}
  \varphi::= \bot \mid c\mid \varphi \land \varphi \mid  \varphi \lor \varphi \mid  \varphi \rightarrow \varphi \quad\text{ where }c\in\mathcal{C}.
\end{align*}
We define $\top$ as $\bot \rightarrow \bot$ and $\neg\varphi$ as $\varphi \rightarrow \bot$ for every formula~$\varphi$.
By $\var(\varphi)$ we denote the set of all variables occurring in all constraint atoms in formula~$\varphi$.
A \emph{theory} is a set of formulas.

An \emph{interpretation} over $\mathcal{X},\mathcal{D}$ is a pair $\langle h,t \rangle$
of valuations over $\mathcal{X},\mathcal{D}$ such that $h\subseteq t$.
The interpretation is \emph{total} if $h=t$.
%
%
\begin{definition}\label{def:satisfaction}
  Given a denotation $\den{\cdot}$,
  an interpretation $\langle h,t \rangle$ \emph{satisfies} a formula~$\varphi$,
  written $\langle h,t \rangle \models \varphi$,
  if 
  \begin{enumerate}
  \item $\langle h,t \rangle \not\models \bot$
  \item $\langle h,t \rangle \models c \text{ if } h\in \den{c}$ \label{item:htc:atom}
  \item $\langle h,t\rangle \models \varphi \land \psi \text{ if }  \langle h,t\rangle \models \varphi \text{ and }  \langle h,t\rangle \models \psi$
  \item $\langle h,t\rangle \models \varphi \lor \psi \text{ if }  \langle h,t\rangle \models \varphi \text{ or }  \langle h,t\rangle \models \psi$
  \item $\langle h,t\rangle \models \varphi \rightarrow \psi
    \text{ if }\langle w,t\rangle \not\models \varphi \text{ or } \langle w,t\rangle \models \psi
    \text{ for }w\in\{h,t\}$
  \end{enumerate}
\end{definition}
%
\noindent
For compactness, we sometimes write $\tuple{t,t}\models\varphi$ simply as $t\models\varphi$.
In the rest of the paper,
we assume a fixed underlying denotation.

A formula $\varphi$ is a \emph{tautology} when~\mbox{$\langle h,t \rangle\models\varphi$} for every interpretation~\mbox{$\langle h,t \rangle$}
(wrt to some underlying denotation).
Hence, a constraint atom $c$ is tautologous whenever \mbox{$\den{c}=\mathcal{V}$}.
We say that an interpretation~$\tuple{h,t}$ is a model of a theory~$\Gamma$,
written $\tuple{h,t} \models \Gamma$,
when $\tuple{h,t} \models \varphi$ for every $\varphi \in \Gamma$.
We write $\Gamma \equiv \Gamma'$ if $\Gamma$ and $\Gamma'$ have the same models.
We omit braces whenever $\Gamma$ (resp.~$\Gamma'$) is a singleton.

A (total) interpretation $\langle t,t\rangle$ is an \emph{equilibrium model} of a theory~$\Gamma$,
if $\langle t,t\rangle \models \Gamma$ and there is no $h\subset t$
such that $\langle h,t\rangle \models \Gamma$.
Valuation~$t$ is also called a \emph{stable model} of $\Gamma$ and $\mathit{SM}(\Gamma)$ collects the set of stable models of~$\Gamma$.

Finally,
constraint atoms also allow us to capture constructs similar to aggregates.
For instance,
a constraint atom
\begin{gather}
  \mathit{sum}\agg{ \ s_1,s_2,\dots \ } = s_0
  \label{eq:sum}
\end{gather}
with each $s_i\in\mathcal{X} \cup \mathcal{D}$ for $0\leq i$
can express that the (possibly infinite) sum of the values associated with the expressions $s_1, s_2, \dotsc$ is equal to~$s_0$.
The semantics of this constraint atom can be given by the following denotation:
\begin{align*}
  \den{ &\mathit{sum}\agg{ s_1,s_2,\dots } = s_0 } \ =\\
        &\qquad\{ v\in\mathcal{V}  \mid v(s_i)\in\mathbb{Z},i\geq 1, v(s_0)=\textstyle{\sum_{i\geq 1} v(s_i)} \}
\end{align*}
This kind of construct allows us to gather the total tax revenue of the country
with an expression of the form~$\mathit{sum}\agg{ \mathit{tax}(p_1) ,\mathit{tax}(p_2) ,\dots }$
where $p_1,p_2$ are all the people in the country.
We abbreviate this expression as~$\mathit{sum}\agg{ \, \mathit{tax}(P) \,  }$.
Note however that such simple aggregate-like constructs do not allow for obtaining the total tax revenue of each region $R$,
as in~\eqref{eq:tax.sum}.
What is missing is the possibility of applying the $\mathit{sum}$ operation to a set of conditional expressions.
We address this issue in the next section.
%

%
\section{Extending \HTC\ with Conditional Constraints}
\label{sec:ht:cc}

We now extend the logic $\mathit{HT}_{\!C}$ with conditional expressions,
inspired by the concept of aggregate elements in ASP~\cite{aspcore2}.
While atoms are naturally conditioned by using an implication,
a formal account is needed for conditioning subatomic expressions.

As before,
we leave the syntax of expressions open as arbitrary strings, possibly combining elements from $\mathcal{X}$ and $\mathcal{D}_\undefined$,
but we additionally assume now that some subexpressions, called \emph{conditional expressions}, may have the form \cterm{s}{s'}{\varphi} where $\varphi$ is a formula called the \emph{condition}.
An expression or constraint atom which, in its turn, does not contain any conditional expression is called \emph{condition-free}.
We do not allow nested conditional expressions, that is,
we assume that $s$ and $s'$ are condition-free expressions and that constraint atoms in formula $\varphi$ are also {condition-free}.
The intuitive reading of \cterm{s}{s'}{\varphi} is ``get the value of $s$ if $\varphi$, or the value of $s'$ otherwise.''
According to this reading,
we must establish some connection among subexpressions \cterm{s}{s'}{\varphi}, $s$ and $s'$ both in the possible constraints $\mathcal{C}$
we can form and in the interrelation among their valuations.
Therefore, if we have a constraint atom in $\mathcal{C}$ that has form
\(
c[\tau]
\)
with $\tau=\cterm{s}{s'}{\varphi}$, we require that the constraint atoms
\(
c[\tau/s],c[\tau/s'],c[\tau/\undefined]
\)
also belong to~$\mathcal{C}$.
Moreover, the valuations for these constraint atoms must satisfy
\begin{enumerate}[start = 4]
\item $v \in \den{ \ c[\tau/\undefined ] \ }$ implies
  $v \in \den{ \ c[\tau/s] \ }$
  and
  $v \in \den{ \ c[\tau/s'] \ }$
  \label{den:prt:3}
\end{enumerate}
Condition~\ref{den:prt:3} strengthens Condition~\ref{den:prt:0} for the case of conditional constraint atoms.
Intuitively,
it says that, if a constraint does not hold for some subexpression, then it cannot hold when that subexpression is left undefined.
For instance, if we include a constraint atom $x-\cterm{y}{z}{p} \leq 4$,
then we must allow for forming the three constraint atoms $x-y \leq 4$ and $x-z \leq 4$ and $x-\undefined \leq 4$, too,
and any valuation for the latter must also be a valuation for the former two.

Satisfaction of constraint atoms is defined by a previous syntactic unfolding of their conditional subexpressions,
using some interpretation~$\langle h,t \rangle$ to decide the truth values of formulas in conditions.
%
\begin{definition}
  Given an interpretation $\langle h,t \rangle$ and a conditional expression $\tau=(\ctermm{s}{s'}{\varphi})$ we define:
  \begin{align}
    \label{def:eval:term}
    \eval{h}{t}(\tau)
    &=
      \left\{
      \begin{array}{ll}
        s & \text{if } \langle h,t\rangle\models\varphi\\
        s' & \text{if } \langle t,t\rangle \hspace{2pt}\not\models\varphi\\
        \undefined & \text{otherwise}
      \end{array}
      \right.
  \end{align}
\end{definition}
%
\noindent
Note that the cases for $\langle h,t\rangle\models\varphi$ and $\langle t,t\rangle\not\models\varphi$ agree with our stated intuition that \cterm{s}{s'}{\varphi} gets the value of $s$ if~$\varphi$ is satisfied, or the value of~$s'$ when~$\varphi$ is not satisfied.
The remaining case leaves the expression undefined when
neither $\langle h,t\rangle\models\varphi$
nor $\langle t,t\rangle\not\models\varphi$
hold.

For a constraint atom $c \in \mathcal{C}$,
we define $\eval{h}{t}(c)$ as the constraint atom that results from replacing each conditional expression~$\tau$ in $c$ by $\eval{h}{t}(\tau)$.
Accordingly, $\eval{h}{t}(c)$ is condition-free.
%
As an example, consider the conditional difference constraint
\begin{gather}\label{eq:diff.constraint.conditional}
 x-\cterm{y}{3}{p} \leq 4
\end{gather}
and the valuation ${t=\{ (x,7), (y,0) \}}$.
Then, the result of its evaluation $\eval{t}{t}( x-\cterm{y}{3}{p} \leq 4 )$ is the condition-free difference constraint
$x-3 \leq 4$.
Note that $p$ is not satisfied by $\tuple{t,t}$ and, thus, the conditional expression is replaced by its ``else'' part, viz.~$3$.

Satisfaction of formulas containing conditional terms is then naturally defined by
replacing Condition~\ref{item:htc:atom} in Definition~\ref{def:satisfaction} by:
\begin{enumerate}
\item [\ref{item:htc:atom}$'$\!.] $\langle h,t \rangle \models c \text{ if } h\in \den{\eval{h}{t}(c)}$
\end{enumerate}
In our running example,
we obtain
$\tuple{t,t} \models x-\cterm{y}{3}{p} \leq 4$
because $\tuple{t,t} \models x-3 \leq 4$.

Recall that, due to Condition~\ref{den:prt:0} of denotations,
condition-free constraint atoms behave monotonically,
that is, ${t \subseteq t'}$ and ${\tuple{t,t} \models c}$ imply ${\tuple{t',t'} \models c}$.
However, this no longer holds for conditional constraint atoms, which may behave non-monotonically.
For instance, in our running example,
the valuation ${t' = \{ (x,7), (y,0) , (p, \mathbf{t}) \}}$
satisfies both ${t \subseteq t'}$
and
\mbox{$\tuple{t',t'} \not\models x-\cterm{y}{3}{p} \leq 4$}.
This is because
\mbox{$\eval{t'}{t'}( x-\cterm{y}{3}{p} \leq 4 )$}
yields a condition-free constraint atom different from the one above,
namely $x-y \leq 4$.
This constraint atom is not satisfied by $\tuple{t',t'}$.

The following proposition tells us that usual properties of Here-and-There are still valid in this new extension.%
\footnote{An extended version of the paper including all proofs can be found here: \url{http://arxiv.org/abs/2002.06911}}
Given any \HT\ formula $\varphi$ let $\varphi[\overline{a}/\overline{\alpha}]$ denote the uniform replacement of any tuple of atoms $\overline{a}=(a_1,\dots,a_n)$ in $\varphi$ by a tuple of arbitrary \HTC\ formulas $\overline\alpha = (\alpha_1,\dots,\alpha_n)$.
%
\begin{proposition}\label{prop:htcc:persistence}
  Let $\langle h,t\rangle$ and $\langle t,t\rangle$ be two interpretations,
  and $\varphi$ be a formula. Then,
  \begin{enumerate}
  \item $\langle h,t\rangle \models \varphi$ implies $\langle t,t\rangle \models \varphi$\ ,
    \label{item:1:prop:htcc:persistence}
  \item $\langle h,t\rangle \models \varphi\rightarrow\bot$ iff $\langle t,t\rangle \not\models \varphi$\ ,
    \label{item:2:prop:htcc:persistence}
  \item If $\varphi$ is an \HT\ tautology then $\varphi[\overline{a}/\overline{\alpha}]$ is an \HTC\ tautology.
    \label{item:3:prop:htcc:persistence}
  \end{enumerate}
\end{proposition}

As an example of property \ref{item:3:prop:htcc:persistence} in Proposition~\ref{prop:htcc:persistence}, we can conclude, for instance, that $( x-\cterm{y}{3}{p} \leq 4 ) \to \neg \neg ( x-\cterm{y}{3}{p} \leq 4 )$ is an \HTC\ tautology because we can replace $a$ in the \HT\ tautology $a \to \neg \neg a$ by the \HTC\ formula $( x-\cterm{y}{3}{p} \leq 4 )$.
%
In particular,
the third statement guarantees that all equivalent rewritings in~$\HT$ are also applicable to~$\HTC$.
Note that
every~$\HTC$ theory can be considered as an~$\HT$ theory where each constraint atom is regarded as a proposition without further structure.
As a result, the deductions made in~$\HT$ about a theory are sound with respect to~$\HTC$,
even they may not be complete because~$\HT$ misses the relation between atoms that derive from their internal structure.

\section{Conditional linear constraints} 
\label{sec:lcprograms}
We now focus on constraint atoms for dealing with
\emph{conditional linear constraints} on integer variables, or \emph{linear constraints} for short.
These constraints can be seen as a generalisation of regular aggregate atoms used in ASP.
The syntax of linear constraints is defined as follows:
\begin{gather*}
   \lambda  \ ::= \  d \ \grsep \ d \cdot x \qquad\qquad
\tau  \ ::= \  \lambda \ \grsep \ \cterm{\lambda}{\lambda'}{\varphi}
\end{gather*}
where ${d \in \mathbb{Z} \subseteq \mathcal{D}}$ is an integer constant,
${x \in \mathcal{X}}$ a constraint variable and
$\varphi$ a formula.
We call $\tau$ a \emph{term};
it is either a \emph{linear term}, $\lambda$, or
a \emph{conditional term} of form \cterm{\lambda}{\lambda'}{\varphi}.
A \emph{linear expression} $\alpha$ is a possibly infinite sum $\tau_1 + \tau_2 + \dots$ of terms $\tau_i$.
Then, a \emph{linear constraint} is an inequality \mbox{$\alpha \leq \beta$} of linear expressions~$\alpha$ and~$\beta$.
We denote the set of variables occurring in $\alpha$ by $\vars{\alpha}$.
%
A linear constraint
\mbox{$\alpha \leq \beta$} is said to be in normal form if $\beta=d \in \mathbb{Z}$.
We adopt some usual abbreviations.
We simply write $x$ instead of $1\cdot x$ and we directly replace the `$+$' symbol by (binary) `$-$' for negative constants.
Moreover, when clear from the context, we sometimes omit the `$\cdot$' symbol
and parentheses.
We do not remove parentheses around conditional expressions.
As an example,  $-x + \ \cterm{3y}{2y}{\varphi} \ - 2 z$ stands for
$(-1) \cdot x + \cterm{3 \cdot y}{2 \cdot y}{\varphi} + (-2) \cdot z$.
Other abbreviations must be handled with care.
In particular, we neither remove products of form $0 \cdot x$ nor replace them by $0$ (this is because $x$ may be undefined, making the product undefined, too).

We also extend the $\leq$ symbol in \mbox{$\alpha \leq \beta$} to other comparison relations defined as the following abbreviations of formulas:
$(\alpha < \beta) \eqdef \alpha \leq \beta \wedge \neg (\beta \leq \alpha)$, \
$(\alpha=\beta) \eqdef (\alpha \leq \beta) \wedge (\beta \leq \alpha)$ and
$(\alpha \neq \beta) \eqdef (\alpha < \beta) \vee (\beta < \alpha)$.
Notice that \mbox{$\alpha \neq \beta$} is stronger than $\neg(\alpha = \beta)$ since the former requires $\alpha$ and $\beta$ to have different values (and so, to be both defined), while the latter checks that $\alpha=\beta$ does not hold, and this includes the case in which any of the two is undefined.
For any linear expression~$\alpha$, we define $\df{\alpha} \eqdef \alpha \leq \alpha$ to stand for ``$\alpha$~is defined,'' that is, $\alpha$ has a value.

For the semantics of linear constraints ${\alpha \leq \beta}$, we resort to~\cite{cakaossc16a} where $\alpha$ and $\beta$ were \emph{condition-free}, that is, they were sums of linear terms.
As shown there, given any linear expression $\alpha=\lambda_1+\lambda_2+\dots$ of that form, we can define their partial valuation $\val$ so that $\val(\alpha)$ corresponds to:
\begin{gather*}
  \val(\lambda_1 + \lambda_2 + \dots )
  \eqdef
  \left\{
    \begin{array}{cl}
      \undefined & \text{if } \val(\lambda_i) \notin \mathbb{Z},
                 \text{for some } \lambda_i\\[5pt]
      \sum_{i\geq 0} \val(\lambda_i) & \text{otherwise}
    \end{array}
  \right.
\end{gather*}
where $\val(\lambda)$ for non-constant linear terms is defined as expected:
$\val(k \cdot x) \eqdef k \cdot \val(x)$ if $\val(x) \in \mathbb{Z}$,
and $\undefined$ otherwise.
In other words, a (condition-free) linear expression is evaluated as usual, except that it is undefined if it contains some undefined subterm (or eventually, some undefined variable).
Then, the denotation of a condition-free linear constraint $\alpha \leq \beta$ is defined as:
\begin{gather*}
\den{\alpha \leq \beta} \quad\eqdef\quad \{\val \mid \val(\alpha), \val(\beta)\in \mathbb{Z}, \val(\alpha)\leq \val(\beta)\}
\end{gather*}
These are valuations $v$ in which \mbox{$\val(\alpha)\leq \val(\beta)$} holds as expected,
but both values $\val(\alpha)$ and $\val(\beta)$ must be defined integers.
It is easy to see that, when $\alpha \leq \beta$ is condition-free,
it can only be satisfied at $h$, if all variables occurring in the constraint are defined in $h$.

When we move to evaluating conditional terms,
we further need some interpretation $\tuple{h,t}$ to decide the satisfaction of formulas in conditions.
The following result asserts that, if $h$ assigns some value to a term $\tau$ (conditional or not),
this value is also preserved in $t$.
\begin{proposition}\label{prop:arithmetic:persistence}
For any term $\tau$ and interpretation $\tuple{h,t}$,
if $h(\eval{h}{t}(\tau))\neq \undefined$,
then $h(\eval{t}{t}(\tau))=t(\eval{h}{t}(\tau))$.
\end{proposition}

Now, for satisfaction of a conditional linear constraint, it suffices to apply Condition~\ref{item:htc:atom}$'$ of the previous section:
\begin{gather*}
\tuple{h,t} \models \alpha \leq \beta \quad\text{ if }\qquad h \in \den{\eval{h}{t}(\alpha \leq \beta)}
\end{gather*}
That is, we remove conditional terms by applying
\mbox{$\eval{h}{t}(\alpha \leq \beta)$} and then use the denotation of the resulting condition-free linear constraint.

An important consequence of the introduction of conditional terms is that a linear constraint
\mbox{$\alpha \leq \beta$} may now be satisfied even though some of its variables are undefined.
For instance, the constraint
\begin{gather}
    x+y > 1
    \label{eq:x+y}
\end{gather}
is not satisfied for interpretation $t=\{(y,5)\}$ since $t(x)=\undefined$ and we cannot compute $\undefined+5$.
However, the conditional linear expression
\begin{gather}
    \cterm{x}{0}{\df{x}} + \cterm{y}{0}{\df{y}} > 1
    \label{eq:sum.x+y.linear}
\end{gather}
checks whether $x+y$ is greater than one, but replaces any of these two variables by $0$ when they are not defined.
Take the example $t = \{ (y,5) \}$ where $x$ is undefined and $y$ is $5$.
Then, the result of applying function $\eval{t}{t}$ to~\eqref{eq:sum.x+y.linear} amounts to $0+y>1$,
which is satisfied by $t$, that is, $t \in \den{0+y>1}$.



Linear constraints offer a comfortable setting for a practical implementation of the $\mathit{sum}$ aggregate we informally presented in the introduction, since their computation can be eventually delegated to a specialized constraint solver, as done in~\cite{cakaossc16a}.
Besides, other common aggregates such as $\mathit{count}$, $\mathit{max}$ and $\mathit{min}$ can be defined relying on $\mathit{sum}$, as we show below.
A straightforward encoding of~\eqref{eq:sum} is the linear constraint
\begin{align}
    s_1 + s_2 + \dotsc = s_0
    \label{def:sum.as.linear}
\end{align}
which abbreviates the conjunction of the following two linear constraints in normal form:
\begin{eqnarray}
    (- s_0 + s_1 + s_2 + \dots  \leq 0) \label{eq:sum.lp.normal.form1}\\
    (s_0 - s_1 - s_2 - \dots  \leq 0) \label{eq:sum.lp.normal.form2}
\end{eqnarray}
Actually, this encoding allows us to extend the expressions $s_i$ in the aggregate to be any linear or conditional term now.
For the case of variables and constants $s_i \in \mathcal{X} \cup \mathcal{D}$, it is easy to check that the previous denotation we defined for $\sumf$, $\den{ \eqref{eq:sum} }$, is equal to $\den{\eqref{eq:sum.lp.normal.form1}} \cap \den{\eqref{eq:sum.lp.normal.form2}}$ and this, in its turn, means that $\tuple{h,t} \models \eqref{eq:sum}$ iff $\tuple{h,t} \models \eqref{eq:sum.lp.normal.form1} \wedge \eqref{eq:sum.lp.normal.form2}$, i.e. \eqref{eq:sum} and~\eqref{def:sum.as.linear} are logically equivalent.
The encoding of $\sumf$ in~\eqref{def:sum.as.linear} also clarifies its multiset behavior.
For instance, atom
\begin{gather}
    \htagg{sum}{ \ x , \ y \ }{>}{1}
    \label{eq:sum.x+y}
\end{gather}
amounts now to \eqref{eq:x+y}, that is, $x+y>1$ and there is no problem for collecting several occurrences of the same value as in interpretation $\{ (x,1), (y,1) \}$ since $\htagg{sum}{1 , 1}{>}{1}$ amounts to $1+1>1$ which is obviously true.

Up to now, using conditional terms inside $\sumf$ allows us writing, for instance, 
\begin{gather*}
\htagg{sum}{\ \cterm{x}{-x}{x \geq 0} \ , \ y \ }{>}{1}
\end{gather*}
to replace $x$ in \eqref{eq:sum.x+y} by its absolute value, leading to the linear constraint $\cterm{x}{-x}{x \geq 0}+y >1$.
A more common situation, however, is to use a condition to decide whether a term should be included in the multiset or not.
To this aim, we \emph{redefine} the $\sumf$ construct so that its syntax follows the general pattern
\begin{align}
    \htag{sum}{ \ \lambda_1\!:\!\varphi_1, \ \lambda_2\!:\!\varphi_2, \ \dots \ }
    \label{def:sum.conditions}
\end{align}
where $\lambda_i$ are linear terms and $\varphi_i$ are condition-free formulas.
Semantically,
we assume that the new notation \eqref{def:sum.conditions} is just an abbreviation of the linear expression $\theta_1+\theta_2+\dots$ where each $\theta_i$ corresponds now to the conditional term:
\begin{gather*}
\cterm{\lambda_i}{0}{\ \varphi_i \wedge \df{\lambda_i}\ }
\end{gather*}
Note first that $\theta_i$ becomes $0$ when its condition is not fulfilled,
being a simple way to remove $\lambda_i$ from the multiset of the sum.
A second important observation is that we have reinforced the condition~$\varphi_i$ with the formula $\df{\lambda_i}$ so that, if the term $\lambda_i$ is undefined, the conditional also becomes $0$ rather than $\undefined$.
This behavior is interesting since some sums may be formed using undefined variables,
either because no information has been provided for them,
or because they come from unwanted, undefined terms generated from grounding, such as, say $tax(3/0)$.

When $\varphi_i=\top$, we allow to replace the multiset element ``$\lambda_i:\top$'' by ``$\lambda_i$''.
As an example, notice that, under this new understanding, the aggregate atom \eqref{eq:sum.x+y} corresponds now to:
\begin{gather*}
\htag{sum}{ \ x\!:\!\top, \ y\!:\!\top \ } > 1
\end{gather*}
and its translation into a linear constraint eventually corresponds to~\eqref{eq:sum.x+y.linear} rather than \eqref{eq:x+y}.
Thus, the atom may still be true even though some variable is undefined, as we discussed before.
In fact, it is not difficult to see that an aggregate expression $\alpha$ like \eqref{def:sum.conditions} is
always defined for any valuation $v$
because $v(\eval{v}{v}(\alpha))\in \mathbb{Z}$.
Note that an aggregate expression may still be undefined for interpretations~$\tuple{h,t}$ due to a different evaluation of one of its conditions between~$h$ and~$t$.
We assume this understanding from now on.
%

\section{Programs with linear constraints and aggregates}

In this section, we provide a logic programming language based on a syntactic fragment of $\HTC$.
In principle, logic programming rules can be built as usual, that is, implications (usually written backwards) $\mathit{Head} \leftarrow \mathit{Body}$ where $\mathit{Head}$ is a disjunction of atoms and $\mathit{Body}$ a conjunction of literals (that is, atoms or their default negation).
However, a constraint atom in the head must be handled with care, since it does not provide any directionality for its set of variables.
For instance, if we want that $\mathit{tot}$ gets the value of some variable~$x$, we cannot just use the rule
$
(\mathit{tot}=x \leftarrow \top )
$
because there is no difference wrt.\ \mbox{$x=\mathit{tot}$}.
In fact, without further information in the program, this rule would assign some arbitrary value for both~$x$ and $\mathit{tot}$ to make the constraint atom true.
To allow for directional assignments, \cite{cakaossc16a} introduced the following construct.
An \emph{assignment}~$A$ for variable $x$ is an expression of the form $\ass{x}{\alpha}{\beta}$ (with $\alpha,\beta$ linear expressions) standing for the formula
\begin{align}
\neg \neg \df{A} \wedge (\df{A} \rightarrow \alpha \leq x \wedge x \leq \beta) \label{f:assig}
\end{align}
where $\df{A} \eqdef \df{\alpha} \wedge \df{\beta}$.
An assignment~$A$ is \emph{applicable} in $\tuple{h,t}$ when $\tuple{h,t} \models \df{A}$.
The \mbox{non-directional} version of assignment $A$ is defined as
\mbox{$\Phi(\ass{x}{\alpha}{\beta}) \,\eqdef\, (\alpha \leq x) \wedge (x \leq \beta)$}.
We see that an assignment~$A$ makes some additional checks regarding the definedness of $\alpha$ and $\beta$
before imposing any condition on the variable~$x$.
In particular, $(\df{A} \rightarrow \alpha \leq x \wedge x \leq \beta)$ guarantees that $\alpha$ and $\beta$ can be used to fix the value of $x$, but not of variables in $\alpha$ and $\beta$ themselves.
On the other hand, $\neg \neg \df{A}$ can be seen as a constraint checking that $\alpha$ and $\beta$ must be eventually defined in the stable model, but through other rule(s) in the program.
When the upper and lower bounds coincide, we just write $(x := \alpha) \eqdef (\ass{x}{\alpha}{\alpha})$, that is, $\neg \neg \df{\alpha} \wedge (\df{\alpha} \rightarrow x=\alpha)$.
As a result, $\Phi(x:=\alpha)=(x=\alpha)$.
The following proposition relates an assignment~$A$ and its non-directional version~$\Phi(A)$
in some particular interesting cases.
\begin{proposition}\label{prop:assign:nondirect}
Given an assignment $A = (\ass{x}{\alpha}{\beta})$, we have
\begin{enumerate}
\item $A \wedge \df{A} \equiv \Phi(A)$
\item $\neg A \equiv \neg \Phi(A)$
\end{enumerate}
\end{proposition}
\noindent
In particular, if $A=(\ass{x}{\alpha}{\beta})$ contains no variables other than the assigned $x$, then $\df{A}=\top$ and so $A \equiv \Phi(A)$.

We are now ready to introduce the syntactic class of logic programs.
A \emph{linear constraint rule}, or \LC-rule for short, is a rule of the form:
\begin{align}
A_1 ; \dots ; A_n \leftarrow B_1, \dots, B_m, \neg B_{m+1}, \dots, \neg B_k \label{f:rule}
\end{align}
\noindent with \mbox{$n\geq 0$} and \mbox{$k\geq m\geq 0$},
where each $A_i$ is an \emph{assignment}
and each $B_j$ is a \emph{linear constraint}.
For any rule $r$ like \eqref{f:rule},
we let $\Head(r)$ stand for the set $\{A_1,\dots,A_n\}$
and $\Body(r)$ be the set $\{B_1, \dots, B_m, \neg B_{m+1}, \dots, \neg B_k\}$.
By abuse of notation, we sometimes use~$\Head(r)$
to stand for the disjunction~$\bigvee \Head(r)$
and $\Body(r)$ to stand for the conjunction~$\bigwedge \Body(r)$.
An $\HTC$ theory consisting of \LC-rules only is called \emph{\LC-program}.

As an example,
the following \LC-rule corresponds to one of the ground instances of the rule~\eqref{eq:tax.sum} in the introduction.
\begin{align}
\mathit{total}(r) := \alpha \ \leftarrow \ \mathit{region}(r)
	\label{f:ex1z}
\end{align}
with
\mbox{$\alpha = \mathit{sum}\agg{ \, \mathit{tax}(p_1) \!:\! \mathit{lives}(p_1,r),   \mathit{tax}(p_2) \!:\! \mathit{lives}(p_2,r), \dotsc  \,  }$}.
Intuitively,
rule~\eqref{f:ex1z} states that the value of
$\mathit{total}(r)$ is equal to the value computed by $\alpha$ whenever the Boolean variable~$\mathit{region}(r)$ holds.
This rule says nothing about the value of
$\mathit{total}(r)$ if the condition $\mathit{region}(r)$ does not hold.
As a result, in this case, and in absence of other rules defining its value, the value of
$\mathit{total}(r)$ should be left undefined.
This has some analogy to the fact that every true atom in a stable model must be supported by some rule in the program.
Similarly, we may expect that every \emph{defined variable} in a stable model is also supported by some \LC-rule.
This result is not immediately obvious, since we allow assignments with conditional linear expressions in the head (which includes aggregate expressions).
We lift next the notion of supported model from standard ASP to the case of \mbox{$\LC$-programs}.
Given a valuation~$v$, a variable $x \in \X$ and an \mbox{$\LC$-program}~$\Pi$, we say that value $d \in \D$ is \emph{supported} for $x$ wrt.\ $\Pi$ and $v$
if there is a rule~$r \in \Pi$ and an assignment of the form~$x := \alpha..\beta$
in the head of $r$ satisfying the following conditions:
\begin{enumerate}
\item $v(x)=d$ and $v(\alpha)\leq d \leq  v(\beta)$, so both $v(\alpha), v(\beta) \in \mathbb{Z}$
\item $v \not\models A'$ for every assignment of the form
$y:=\alpha'..\beta'$ in the head of~$r$ where $y$ is a variable different from~$x$,

\item $v \models \Body(r)$.
\end{enumerate}
We say that a valuation~$v$ is \emph{supported} wrt.\ some \mbox{\LC-program}~$\Pi$
if every $v(x)\neq \undefined$ is a supported value for $x$ wrt.~$\Pi$ and~$v$.

\begin{proposition}\label{prop:lc.supported}
Every stable model of any~$\LC$-program is also supported.
\end{proposition}

Notice that an \LC-rule may contain nested implications in the head due to the presence of assignments.
The following theorem shows that \LC-rules can be unfolded into a set of implications where the antecedent is
a conjunction of literals and the consequent is a disjunction of constraint atoms.
More formally,
an \HTC-rule is an expression of the form of~\eqref{f:rule}
with \mbox{$n\geq 0$} and \mbox{$k\geq m\geq 0$},
where each $A_i$ and $B_j$ is a \emph{linear constraint}.
Note that the difference between \LC- and \HTC-rules resides in the fact that the head of the former are build of assignments while the head of the later are build of linear constraints.

\begin{thm}\label{th:assign}
A rule $r$ as in \eqref{f:rule} is equivalent to the conjunction $\bigwedge_{\Delta \subseteq \Head(r)} \Psi_{\Delta}$
where $\Psi_\Delta$ is the following implication:
\begin{align*}\textstyle
\bigvee_{A \in \Delta}
\Phi(A)
\ \leftarrow \
\Body(r)
\wedge
\bigwedge_{A \in \Delta} \df{A}
\wedge
\bigwedge_{A' \in \Head(r) \setminus \Delta}
\neg \Phi(A')
\end{align*}
\end{thm}

The implication above is not an \HTC-rule yet:
note that each $\Phi(A)$ in the head may be a conjunction of the form \mbox{$\alpha \leq x \wedge x \leq \beta$} and
each $\neg \Phi(A')$ in the body can be a negated conjunction of a similar form that,
by De~Morgan laws, becomes a disjunction \mbox{$\neg (\alpha \leq x) \vee \neg (x \leq \beta)$}.
Still, these constructs can be easily unfolded in \HT\ by distributivity properties that guarantee that
$\varphi \wedge \varphi' \leftarrow \psi$ is equivalent to the pair of rules
$\varphi \leftarrow \psi $ and $\varphi' \leftarrow \psi$ and something analogous for disjunctions in the body.
Therefore, every \mbox{\LC-rule} can be rewritten as a set of \mbox{\HTC-rules}.
As a small illustration, take the \LC-rule~\eqref{f:ex1z} with a single head assignment $A=(\mathit{total}(r) := \alpha)$.
We can only form two sets $\Delta_1=\{A\}$ and $\Delta_2=\emptyset$ that, according to Theorem~\ref{th:assign},
generate the respective implications:
\begin{align}
\mathit{total}(r) = \alpha \ &\leftarrow \ \mathit{region}(r) \wedge \df{\alpha} \label{f:ex1a} \\
\bot & \leftarrow \mathit{region}(r) \wedge \neg (\mathit{total}(r) = \alpha ) \label{f:ex1b}
\end{align}
Moreover, since the aggregate satisfies $t(\eval{t}{t}(\alpha)) \neq \undefined$ for any valuation $t$, we can prove that $\tuple{t,t} \models \df{\alpha}$ and so, by Proposition~\ref{prop:htcc:persistence}, constraint \eqref{f:ex1b} can be equivalently transformed into
\begin{align*}
\bot & \leftarrow \mathit{region}(r) \wedge \neg (\mathit{total}(r) = \alpha ) \wedge \df{\alpha}
\end{align*}
that is an \HT\ consequence of \eqref{f:ex1a}, and so, can be eventually removed.


\section{Implementation Outline} 
\label{sec:translation}

In this section,
we propose a method for implementing LC-programs with conditional aggregates that
relies on a syntactic transformation for removing conditional expressions.
This transformation produces a condition-free set of \HTC-rules that can then be solved as in~\cite{cakaossc16a}, using an off-the-shelf CASP solver as a back-end.
In fact, the reduction to conditional-free syntax can be defined for arbitrary \HTC\ theories, not only LC-programs.
Given a conditional expression $\tau = \cterm{s}{s'}{\varphi}$,
we define formula~$\delta(\tau)$ as the conjunction of the following implications
\begin{IEEEeqnarray}{rCl ;C; l}
\phantom{\neg}\varphi &\wedge& \df{s} &\rightarrow& x_\tau = s
	\label{eq:gz.translation.1}
\\
\neg\varphi &\wedge& \df{s'} &\rightarrow& x_\tau = s'
	\label{eq:gz.translation.4}
\\
\phantom{\neg}\varphi &\wedge& \df{x_\tau} &\rightarrow& x_\tau = s
	\label{eq:gz.translation.2}
\\
\neg\varphi &\wedge& \df{x_\tau} &\rightarrow& x_\tau = s'
	\label{eq:gz.translation.5}\\
\IEEEeqnarraymulticol{3}{r}{\df{x_\tau}} &\rightarrow& \varphi  \vee \neg\varphi
	\label{eq:gz.translation.3}
\end{IEEEeqnarray}
where $x_\tau$ is a fresh variable locally occurring in~$\delta(\tau)$.
These implications are used to guarantee that the new auxiliary variable~$x_\tau$ gets exactly the same value as~$\tau$ under any $\tuple{h,t}$ interpretation.
In that way, the conditional expression can be safely replaced by $x_\tau$ in the presence of $\delta(\tau)$.
In particular,~\mbox{\eqref{eq:gz.translation.1} and \eqref{eq:gz.translation.4}} alone suffice to guarantee that $x_\tau$ gets the value of $s$ when $\varphi$ holds, or the value of $s'$ if~$\neg \varphi$ instead.
To illustrate the effect of \mbox{\eqref{eq:gz.translation.1}-\eqref{eq:gz.translation.4}}, suppose we have the formula $\htagg{sum}{x,y}{>1} \ \to \ p$ and the fact $y=5$.
This amounts to the theory:
\begin{gather*}
y = 5
\hspace{1cm}
\cterm{x}{0}{\df{x}} + \cterm{y}{0}{\df{y}} > 1
	 \ \rightarrow \ p
\end{gather*}
whose unique stable model is~$t = \{ (p,\mathbf{t}), (y,5) \}$ where $p$ becomes true even though $x$ has no value.
If we replace, say, $\tau=\cterm{y}{0}{\df{y}}$ by $x_\tau$ we get
\begin{gather*}
y = 5
\hspace{1cm}
\cterm{x}{0}{\df{x}} + x_\tau > 1
	 \ \rightarrow \ p
\end{gather*}
and that \eqref{eq:gz.translation.1} and \eqref{eq:gz.translation.4} respectively correspond to:
\begin{gather*}
\df{y} \rightarrow x_\tau = y \hspace{2cm}
\neg \df{y} \rightarrow x_\tau = 0
\end{gather*}
after minor simplifications.
The resulting theory also has a unique stable model~$t' =  t \, \cup \{ (x_\tau,5) \}$ that
precisely coincides with $t$ when projected on the original set of variables $\{x,y\}$.

We see that~\eqref{eq:gz.translation.1} and~\eqref{eq:gz.translation.4} provide the expected behavior in this case and,
in fact, are enough to cover the translation $\delta(\tau)$ of any conditional expression inside an \mbox{\LC-program}.
This is because defined variables in \mbox{$\LC$-programs} need to be supported and, by construction, $x_\tau$ cannot occur in the left hand side of any assignment.
Hence, the only way in which $x_\tau$ can be defined is because the body of either~\eqref{eq:gz.translation.1} or~\eqref{eq:gz.translation.4} is satisfied.
%

Implications~\eqref{eq:gz.translation.2}-\eqref{eq:gz.translation.3} are additionally required for the translation of arbitrary theories.
Their need is best illustrated when the constraint atom is used as a rule head or a fact, since this may cause some effect on the involved variables.
The formulas~\eqref{eq:gz.translation.2} and \eqref{eq:gz.translation.5} ensure that variables in $s$ or $s'$ take the correct value when $x_\tau$ is defined.
Take, for instance, the theory only containing a conditional constraint atom~$\cterm{y}{0}{\top} = 5$.
This formula is logically equivalent to $y = 5$ and, thus, it has the stable model~$\{ (y,5) \}$.
If we replace it by some $x_\tau$ and only add~\eqref{eq:gz.translation.1} and~\eqref{eq:gz.translation.4}, we get the theory:
\begin{gather*}
x_\tau = 5
\hspace{20pt}
\top \wedge \df{y} \rightarrow x_\tau = y
\hspace{20pt}
\bot \wedge \df{y} \rightarrow x_\tau = 0
\end{gather*}
(where the last formula is tautological) whose unique stable model is~$\{ (x_\tau,5) \}$ with $y$  undefined.
Now, adding~\eqref{eq:gz.translation.2} and \eqref{eq:gz.translation.5} we also get:
\begin{gather*}
\hspace{20pt}
\top \wedge \df{x_\tau} \rightarrow x_\tau = y
\hspace{20pt}
\bot \wedge \df{x_\tau} \rightarrow x_\tau = 0
\end{gather*}
(again, the last formula is a tautology) whose unique stable model is now~$\{ (x_\tau,5) , (y,5) \}$ as expected.
Finally,~\eqref{eq:gz.translation.3} is added to ensure that $x_\tau$ is only defined in an interpretation~$\tuple{h,t}$ if
either
\mbox{$\tuple{h,t} \models \varphi$}
or
\mbox{$\tuple{h,t} \models \neg\varphi$}.
This correspond to the ``otherwise'' case in Definition~\ref{def:eval:term}.
To show its effect, take the example:
\begin{gather*}
\cterm{y}{y}{p} = 5
\hspace{2cm}
\neg p \to \bot
\end{gather*}
This program has a unique stable model $t = \{ (p,\mathbf{t}) , (y,5) \}$.
Interpretation $\tuple{h,t}$ with $h = \emptyset$ is not a model  because
it does not satisfy
$\eval{h}{t}{(\cterm{y}{y}{p}=5})$
since $\cterm{y}{y}{p}$ is evaluated to
$\undefined$
and $\undefined = 5$ is not satisfied.
On the other hand,
\begin{gather*}
\begin{IEEEeqnarraybox}[][t]{l}
x_\tau = 5
\\
\neg p \to \bot
\end{IEEEeqnarraybox}
\hspace{2cm}
\begin{IEEEeqnarraybox}[][t]{rCl ;C; l}
\phantom{\neg} p &\wedge& \df{y} &\to& x_\tau = y
\\
\phantom{\neg} p &\wedge& \df{x_\tau} &\to& x_\tau = y
\end{IEEEeqnarraybox}
\end{gather*}
has no stable model.
Note that
$\tuple{h',t'}$ with $h' = \{  (y,5) , (x_\tau,5)  \}$
and $t' = \{ (p,\mathbf{t}) , (y,5) , (x_\tau,5) \}$
is a model.
This is solved by adding the following rule corresponding to~\eqref{eq:gz.translation.3}
\begin{IEEEeqnarray*}{l}
\df{x_\tau} \to p \vee \neg p
\end{IEEEeqnarray*}
which is not satisfied by~$\tuple{h',t'}$.

Let us now formalize these intuitions.
We assume that if $c[\tau] \in \C$ is a constraint atom
then \mbox{$x_\tau = s$} and
\mbox{$x_\tau = s'$}
and also
\mbox{$c[\tau/x_\tau]$}
are constraint atoms.
We also assume that for every pair of subexpressions~$s,s'$,
if \mbox{$s = s'$} is a constraint atom, so they are \mbox{$s' = s$} and \mbox{$s = s$}
and that
if \mbox{$s = s'$} and
\mbox{$c[s] \in \C$} are a constraint atoms,
then \mbox{$c[s/s'] \in \C$} is also a constraint atom.
In other words,
if two expressions are of a type that can be syntactically compared,
then replacing one expression by the other also results in a syntactically valid expression.
Furthermore, we require that our denotation behaves as expected wrt.\ equality atoms~``$=$'' and
substitutions of subexpressions,
that is, that it satisfies the following property
\begin{enumerate}[start=5]
\item $v \in \den{s = s'}$ implies
$v \in \den{ \ c[s] \ }$ iff $v \in \den{ \ c[s/s'] \ }$
\end{enumerate}
for any expressions $s,s'$ such that~$c[s]$ and $s=s'$ are constraint atoms.
This condition is similar to Property~\ref{den:prt:1} in Section~\ref{sec:ht:c}
but relates equal subexpressions instead of a variable with its held value.
It is easy to see that the denotation for linear constraints discussed in Section~\ref{sec:lcprograms}
does satisfy this property.
We also extend the definition of $\df{s}$ to arbitrary (non-linear) expressions:
if $s$ is an expression which is not linear, then $\df{s}$ is an abbreviation for $s = s$.

Given a theory~$\Gamma$, by $\delta(\Gamma)$,
we denote the theory resulting from
replacing in $\Gamma$ every occurrence of every conditional expression~$\tau$ by a corresponding fresh variable $x_\tau$
and adding to the result of this replacement the formula $\delta(\tau)$
for every conditional expression~$\tau$ occurring in~$\Gamma$.
Furthermore,
given an interpretation~$\tuple{h,t}$, by $\tuple{h,t}_\tau = \tuple{h_\tau,t_\tau}$,
we denote an interpretation that satisfies the following two conditions:
\[%
\begin{array}{rl}
v_\tau(x)\!=\!v(x) \text{ for } x \!\in\! \X \!\setminus\! \{ x_\tau \}
\end{array}
\hspace{5pt}
\begin{array}{c}
v_\tau(x_\tau) \!=\! \begin{cases}
v(s) &\text{if } \tuple{v,t} \!\models\! \varphi
\\
v(s') &\text{if } \tuple{t,t} \hspace{2pt}\!\not\models\! \varphi
\\
\undefined &\text{otherwise}
\end{cases}
\end{array}
\]
with $v \in \{h,t \}$.
It is easy to see the correspondence of $\tuple{h,t}_\tau$ with the~$\mathit{eval}$ function~\eqref{def:eval:term}: it ensures that the value of $x_\tau$ in the valuation~$v_\tau$ is the same as the conditional expression~$\tau$ in~$v$ for $v \in \{h,t \}$.

\begin{observation}\label{obs:translation}
Any interpretation~$\tuple{h,t}$  and conditional expression~$\tau$
satisfy $h_\tau(x_\tau) = h(\eval{h}{t}{(\tau)})$.
\end{observation}

Now we can relate the construction of interpretation~$\tuple{h,t}_\tau$
with the set of implications~$\delta(\tau)$.

\begin{proposition}\label{prop:translation.model.characterisation}
Any conditional expression~$\tau$ and any
model $\tuple{h,t}$ of~$\delta(\tau)$
satisfy
$\tuple{h,t} = \tuple{h,t}_\tau$.
\end{proposition}
In other words,~$\delta(\tau)$
ensures that $x_\tau$ and $\tau$ have the same evaluation in all models of the resulting theory.
Combining Proposition~\ref{prop:translation.model.characterisation}
and Observation~\ref{obs:translation},
we immediately obtain the following result.

\begin{corollary}\label{cor:translation.theory.equiv}
Let $\Gamma$ be some theory
and
$\tau$ be some conditional expression.
Then,
$\Gamma \cup \{ \delta(\tau) \} \equiv \Gamma[\tau/x_\tau] \cup \{ \delta(\tau) \}$.
\end{corollary}

In other words,
replacing $\tau$ by $x_\tau$ has no effect in a theory that contains the set of implications~$\delta(\tau)$.
To finish the formalization of the correspondence between~$\Gamma$ and $\delta(\Gamma)$,
we resort to the notion of
\emph{projected strong equivalence}~\cite{agcafapepevi19a,eitowo05a}.
Given a set of variables  $X$, let
\mbox{$\mathit{SM}(\Gamma)|_X \eqdef \{ v|_X \mid v \in \mathit{SM}(\Gamma)\}$}
as expected.
Two theories~$\Gamma$ and~$\Gamma'$ for alphabet $\mathcal{X}$ are
said to be \emph{strongly equivalent for a projection} onto ${X \subseteq \mathcal{X}}$,
denoted ${\Gamma \equiv_s^X \Gamma'}$,
iff the equality
${
\mathit{SM}(\Gamma \cup \Delta)|_X = \mathit{SM}(\Gamma \cup \Delta')|_X
}$
holds for any theory $\Delta$ over subalphabet $X$.

\begin{proposition}\label{prop:translation.theory.equiv.proj}
Let $\Gamma$ be some theory,
$\tau$ be some conditional expression
and $X = \X \setminus\{ x_\tau \}$.
Then,
$\Gamma \equiv_s^X \Gamma \cup \{ \delta(\tau) \}$.
\end{proposition}

\begin{thm}\label{thm:translation.theory.equiv}
Let $\Gamma$ be some theory,
$\tau$ be some conditional expression
and $X = \X \setminus\{ x_\tau \}$.
Then,
$\Gamma \equiv_s^X \Gamma[\tau/x_\tau] \cup \{ \delta(\tau) \}$.
\end{thm}

Theorem~\ref{thm:translation.theory.equiv}
is a strongly equivalence result and, thus, the replacement can be made independently of the rest of the theory.
Therefore, this confirms that our translation is strongly faithful and modular.
It is also easy to see that this translation is linear in the size of the program.
In fact, we only introduce as many auxiliary variables as conditional expressions exist in the program and two new constraint atoms, namely $x_\tau = s$ and $x_\tau = s'$
for each new variable.
Furthermore,
the only requirement for the underlying CP solver is to allow the condition-free versions of the constraints appearing in our theory plus equality constraint atoms.
These two requirements are satisfied by off-the-self CASP solvers
when our theory deals with linear constraints.
Finally,
note that the result of translation~$\delta$ is a condition-free \mbox{$\HTC$-theory}.
In particular, when applied to an \mbox{$\LC$-program}, the result is a set containing \mbox{condition-free} \mbox{$\LC$-} and \mbox{$\HTC$-rules}.
\mbox{$\LC$-rules} can be translated into \mbox{$\HTC$-implications} as illustrated by Theorem~\ref{th:assign}.
Although, in general, this translation requires exponential space, a polynomial-size variation is possible.
This is achieved by using auxiliary variables in the fashion of~\cite{tseitin68a}.
The resulting condition-free theory can then be translated into CASP by using
further auxiliary Boolean variables to capture when constraints are defined~\cite{cakaossc16a}.

So far, we have only dealt with $\mathit{sum}$ aggregate functions, though ASP systems usually allow for $\mathit{count}$, $\mathit{max}$ and $\mathit{min}$ operations too.
Compiling $\mathit{count}\agg{\varphi_1,  \varphi_2 , \dotsc }$ into sums is easy: we just transform
it into $\mathit{sum}\agg{1 :\varphi_1, 1 : \varphi_2 , \dotsc }$.
The encoding for the $\mathit{min}$ aggregate is more involved:
We replace an expression of the form
\begin{gather}
	\mathit{min}\agg{s_1 :\varphi_1,\, s_2 : \varphi_2 , \dotsc }
	\label{eq:min}
\end{gather}
by a fresh variable~$x_{\mathit{min}}$ and add the formulas:
\begin{align}
\df{x_{\mathit{min}}} &\leftrightarrow
\mathit{count}\agg{\varphi_1 \!\wedge\! \df{\!s_1\!},\, \varphi_2 \!\wedge\! \df{\!s_2\!}, \dotsc } \geq 1
	\label{eq:min.trans2}
\\
\df{x_{\mathit{min}}} &\to \alpha_{\mathit{min}} \wedge \beta_{\mathit{min}}
	\label{eq:min.trans1}
\end{align}
where $\alpha_{\mathit{min}}$ and $\beta_{\mathit{min}}$ are the following respective expressions
\begin{gather*}
\mathit{count}\agg{\varphi_1 \wedge (s_1 \!<\! x_{\mathit{min}}), \varphi_2 \wedge (s_2 \!<\! x_{\mathit{min}}) , \dotsc } \leq 0
\\
\mathit{count}\agg{\varphi_1 \wedge (s_1 \!\leq\! x_{\mathit{min}}),\, \varphi_2 \wedge (s_2 \!\leq\! x_{\mathit{min}}) , \dotsc } \geq 1
\end{gather*}
We can then simply encode
$\mathit{max}\agg{s_1 :\varphi_1, s_2 : \varphi_2 , \dotsc }$ as the expression
$\mathit{min}\agg{-s_1 :\varphi_1, -s_2 : \varphi_2 , \dotsc }$.
Intuitively,~\eqref{eq:min.trans2} requires that $x_{\mathit{min}}$ is defined iff the multiset has at least one value.
As a result, when the aggregate is defined,
$x_{\mathit{min}}$
can take any value.
Expression~$\alpha_{\mathit{min}}$ in~\eqref{eq:min.trans1} ensures that no element in the aggregate  is strictly smaller than $x_{\mathit{min}}$
while~$\beta_{\mathit{min}}$
guarantees that at least one element is smaller or  equal to~$x_{\mathit{min}}$.

This translation is similar to the one for $\mathit{min}$ aggregates in regular ASP~\cite{alfage15a},
but taking care of the definiteness of variables.
It is worth mentioning that in most approaches to aggregates in regular ASP~\cite{fapfle11a,ferraris11a,siniso02a,pedebr07a,sonpon07b},
replacing the aggregate expression by an auxiliary variable results in a non-equivalent formula, while it is safe in our framework.
In this sense, our approach behaves similarly to ASP aggregates as defined in~\cite{cafafape18a,gelzha19a}.
We discuss this relation in more detail in the next Section.

As a simple example,
consider the translation of the expression~$\mathit{min}\agg{x,y}$.
If both~$x$ and~$y$ are undefined,
so it is the right hand side of~\eqref{eq:min.trans2}.
This forces $x_{\mathit{min}}$ to be undefined as well.
Otherwise,~the right hand side of~\eqref{eq:min.trans2} is defined and so it is~$x_{\mathit{min}}$.
In such case,
a valuation~$t$ satisfies the following two formulas:
\begin{gather*}
\mathit{sum}\agg{1 : x < x_{\mathit{min}},\, 1 : y < x_{\mathit{min}}} \leq 0
\\
\mathit{sum}\agg{1 : x \leq x_{\mathit{min}},\, 1 : y \leq x_{\mathit{min}}} \geq 1
\end{gather*}
If either $t(x)$ or $t(y)$ are strictly smaller than $t(x_{\mathit{min}})$,
then the first formula corresponding to $\alpha_{\mathit{min}}$ is violated.
Similarly, if both $t(x)$ and $t(y)$ are strictly greater than $t(x_{\mathit{min}})$,
then $\beta_{\mathit{min}}$ is violated.
Hence,
$t(x_{\mathit{min}})$
is the minimum of $t(x)$ and $t(y)$.
That is, the value of~$x_{\mathit{min}}$ indeed is the minimal value of~$x$ and~$y$
whenever both~$x$ and~$y$ are defined.
Finally, in any interpretation~$\tuple{h,t}$ in which any of the variables is defined in~$t$ but not in~$h$,
we get that that the right hand side of~\eqref{eq:min.trans2} is also defined in~$t$ but not in~$h$.
The same applies to~$x_{\mathit{min}}$ as a result.
This last case happens as a result of a cyclic dependence as, for instance,
in $x = 1 \leftarrow \mathit{min}\agg{x,y} \geq 1$.

\section{Discussion}\label{sec:discussion}

\HTC\ is a logic to capture non-monotonic constraint theories that permits assigning default values to constraint variables.
Since \HT\ and thus also ASP are special cases of this logic,
it provides a uniform framework integrating ASP and CP on the same semantic footing.
We elaborate on this logic by incorporating \emph{aggregate expressions},
one of the essential elements in ASP's modeling language.
This was missing so far.
We accomplished this by introducing the construct of \emph{conditional expressions} that
allow us to consider two alternatives while evaluating constraints.
With it, we can also deal with aggregate expressions on the constraint side.
To the best of our knowledge, this is the first account that
allows for the use of ASP-like aggregate expressions within constraints.
In particular,
we focus on a fragment of~$\HTC$ that constitutes an extension of logic programs with conditional linear constraints, called $\LC$-programs.
We show that $\mathit{sum}$, $\mathit{count}$, $\mathit{max}$ and $\mathit{min}$ aggregate atoms can be regarded as special cases of conditional linear constraints
and, in fact, our formalism permits their use as terms inside linear constraints.

Condition-free~$\HTC$ captures a fragment of ASP with partial functions~\cite{balduccini12a,cabalar11a}
where constraint variables correspond to $0$-ary evaluable functions.
This work was extended with intensional sets in~\cite{cafafape18a},
where it is shown to capture Gelfond-and-Zhang semantics for ASP with aggregate atoms~\cite{gelzha19a}.
Recall that a characteristic feature of this approach is the adherence to the \emph{vicious circle principle},
stating that ``\emph{no object or property may be introduced by a definition that depends on that object or property itself.}''
As a result,
the ASP program consisting of the single rule
\begin{gather}
  p(a) \ \leftarrow \ \mathit{count}\{ \, X:p(X) \, \}\geq 0. \label{f:count1}
\end{gather}
has no stable model under this semantics.
This distinguishes it from other alternatives~\cite{fapfle11a,ferraris11a,siniso02a,pedebr07a,sonpon07b}
that consider $\{ p(a) \}$ as a stable model.
Note that the only rule supporting $p(a)$ depends on a set which contains it as one of its elements.
Thus, in accordance to the above rationality principle, this is rejected.
In our framework, we can write a very simple rule which reflects a similar behavior
\begin{gather}
x:=1 \ \leftarrow \ \mathit{sum}\{ \, x : \top \, \}\geq 0. \label{f:count1.constraint}
\end{gather}
As above,
this theory has no stable model,
showing that our framework adheres to this rationality principle as well.
As happens with the relation between condition-free~$\HTC$ and ASP with partial functions,
we conjecture that our framework captures a fragment of~\cite{cafafape18a}.
As a result, an \emph{instantiation} process similar to the one in~\cite{fapfle11a} would also allow us to capture~\cite{gelzha19a}.
Confirming this conjecture is ongoing work.

Recall that~\cite{gelzha19a} showed that Gelfond-and-Zhang semantics coincides with the other alternatives~\cite{fapfle11a,ferraris11a,sonpon07b} on programs which are \emph{stratified on aggregates}, which is the fragment covered by the ASP~Core~2 semantics~\cite{aspcore2}.
We have also considered the definition of a semantics for constraint aggregates closer to Ferraris'~\cite{ferraris11a} but the implementation for the latter is not so straightforward as the one shown in this paper and is still under study.

Despite the close relation with~\cite{cafafape18a},
a distinctive feature of our approach is its orientation as a general abstraction of a hybrid solver with the ASP solver in charge of evaluating the Boolean part of the theory while relegating the evaluation of constraint atoms to dedicated CP solvers.
Though, we focus here on reasoning with linear constraints,
our formalism can also be regard as an abstraction of a multi-theory solver where the semantics of different constraint atoms are evaluated by different CP solvers.
Interestingly,
we provide a polynomial translation from $\HTC$ with conditional constraints to (condition-free) CASP theories.
This allows us to use off-the-shelf CASP solvers as back-ends for implementing our approach.
This is also ongoing work.
%

\acknowledgements

This work was partially supported by Ministry of Science and Innovation, Spain (TIC2017-84453-P),
Xunta de Galicia, Spain (GPC ED431B 2019/03 and 2016-2019 ED431G/01, CITIC Research Center),
and German Research Foundation, Germany (SCHA 550/11).



\appendix

\clearpage
\section{Proofs of results}
\label{sec:proofs}
\begin{proof}[Proof of Proposition~\ref{prop:htcc:persistence}]
In the following proof sketches for \ref{item:1:prop:htcc:persistence} and \ref{item:2:prop:htcc:persistence}, we focus on Condition 2'.
The other cases are proven in $HT_c$ without conditional expression and the full proof is obtained via structural induction.
\begin{enumerate}
\item[\ref{item:1:prop:htcc:persistence}] $\langle h,t\rangle \models \varphi$ implies $\langle t,t\rangle \models \varphi$
 $\langle h,t\rangle \models c$ implies $\langle t,t\rangle \models c$ for $c\in\mathcal{C}$:
     \begin{align}
     \text{Assume } &\langle h,t\rangle\models c\label{prop:htcc:persistence:eq1}\\
     \Rightarrow\, &h\in \den{\eval{h}{t}(c)}\label{prop:htcc:persistence:eq2}\\
     \Rightarrow\, &t \in \den{\eval{t}{t}(c)}\label{prop:htcc:persistence:eq3}\\
     \Rightarrow\, &\langle t,t\rangle\models c \label{prop:htcc:persistence:eq4}
     \end{align}
    Implication between \eqref{prop:htcc:persistence:eq1} and \eqref{prop:htcc:persistence:eq2} holds by definition of the satisfaction relation.
    Implication between \eqref{prop:htcc:persistence:eq2} and \eqref{prop:htcc:persistence:eq3} holds by definition of the evaluation function and conditions \ref{den:prt:0} and \ref{den:prt:3} for denotations,
          since for any conditional expression $\tau=\cterm{s}{s'}{\varphi}$ in $c$, either
	    \begin{enumerate}
	     \item $\tuple{h,t}\models\varphi$, then by persistence since $\varphi$ is condition-free $\tuple{t,t}\models\varphi$ and therefore $\eval{h}{t}(\ctermm{s}{s'}{\varphi})=\eval{t}{t}(\ctermm{s}{s'}{\varphi})=s$,

	     \item $\tuple{t,t}\not\models\varphi$, and therefore
       $\eval{h}{t}(\ctermm{s}{s'}{\varphi})=\eval{t}{t}(\ctermm{s}{s'}{\varphi})=s'$,

	     \item or $\eval{h}{t}(\ctermm{s}{s'}{\varphi})=\undefined$.
	    \end{enumerate}
	 For (a) and (b) evaluation of $\tau$ is identical.
	 For (c), we have $c[\tau/\undefined]$ when evaluating in $\tuple{h,t}$ 
	 and either $c[\tau/s]$ or $c[\tau/s']$  when evaluating in $\tuple{t,t}$. 
	 In both cases, $h\in\den{c[\tau/\eval{h}{t}(\tau)]}$ implies $h\in\den{c[\tau/\eval{t}{t}(\tau)]}$
	 due to Condition~\ref{den:prt:3}. 
	 Thus, we have $h\in\den{\eval{h}{t}(c)}$ implies $h\in\den{\eval{t}{t}(c)}$,
	 and ultimately $t\in\den{\eval{t}{t}(c)}$ due to Condition~\ref{den:prt:0}.

   Implication between \eqref{prop:htcc:persistence:eq3} and \eqref{prop:htcc:persistence:eq4} holds by definition of the satisfaction relation
\item[\ref{item:2:prop:htcc:persistence}] $\langle h,t\rangle \models \varphi\rightarrow\bot$ iff $\langle t,t\rangle \not\models \varphi$.
  \begin{itemize}
   \item $\langle h,t\rangle \models \varphi\rightarrow\bot$ implies $\langle t,t\rangle \not\models \varphi$ for any formula $\varphi$
     \begin{itemize}
      \item[] $\langle h,t\rangle\models\varphi\rightarrow \bot$
      \item[$\Rightarrow$]  $\langle t,t\rangle\models\varphi\rightarrow \bot$ due to~\ref{item:1:prop:htcc:persistence}
      \item[$\Rightarrow$] $\langle t,t\rangle\not\models\varphi\text{ or }\langle t,t\rangle\models\bot$
      \item[$\Rightarrow$] $\langle t,t\rangle\not\models\varphi$
     \end{itemize}
   \item $\langle t,t\rangle \not\models \varphi$ implies $\langle h,t\rangle \models \varphi\rightarrow\bot$ for $c\in\mathcal{C}$
     \begin{itemize}
      \item[] $\langle t,t\rangle\not\models \varphi$
      \item[$\Rightarrow$] $\langle h,t \rangle\not\models \varphi$ due to~\ref{item:1:prop:htcc:persistence}
      \item[$\Rightarrow$] $\langle h,t\rangle\not\models \varphi\text{ or }\langle h,t\rangle\models\bot\text{ and }\langle t,t\rangle\not\models \varphi\text{ or }\langle t,t\rangle\models\bot$
      \item[$\Rightarrow$] $\langle h,t \rangle\models \varphi\rightarrow\bot$
     \end{itemize}
  \end{itemize}
\end{enumerate}
We define $\At{\tuple{h,t}}=\{c\in \mathcal{C} \mid h\in\den{\eval{h}{t}(c)}\}$ for any interpretation $\tuple{h,t}$.
Then, $\tuple{\At{\tuple{h,t}},\At{\tuple{t,t}}}$ is a valid \HT\ interpretation
due to $\At{\tuple{h,t}}\subseteq\At{\tuple{t,t}}$,
which follows from $c\in\At{\tuple{h,t}}$
then $h\in\den{\eval{h}{t}(c)}$ implying $t\in\den{\eval{t}{t}(c)}$ due to Proposition~\ref{prop:htcc:persistence}.\ref{item:1:prop:htcc:persistence},
and thus $c\in\At{\tuple{t,t}}$.

We proof property~\ref{item:3:prop:htcc:persistence} in Proposition~\ref{prop:htcc:persistence} by proofing $\tuple{h,t}\models\varphi$ iff  $\tuple{\At{\tuple{h,t}},\At{\tuple{t,t}}}\models\varphi$ for any interpretation~$\tuple{h,t}$ and formula~$\varphi$, and thus tautologies are preserved between \HT\ and \HTC.
Since satisfaction relations are identical except for Condition 2', we focus on $\varphi=c$ for $c\in\mathcal{C}$ as the induction base.
The other cases follow by induction since they are identical betewen~\HT\ and \HTC.

We proof $\tuple{h,t}\models c$ iff  $\tuple{\At{\tuple{h,t}},\At{\tuple{t,t}}}\models c$ for $c\in\mathcal{C}$:
     \begin{align}
                   &\langle h,t\rangle\models c\label{prop:htcc:persistence:eq5}\\
     \text{iff }\, &h\in \den{\eval{h}{t}(c)}\label{prop:htcc:persistence:eq6}\\
     \text{iff }\, &c \in \At{\tuple{h,t}}\label{prop:htcc:persistence:eq7}\\
     \text{iff }\, &\tuple{\At{\tuple{h,t}},\At{\tuple{t,t}}}\models c \label{prop:htcc:persistence:eq8}
     \end{align}
Equivalence between \eqref{prop:htcc:persistence:eq5} and \eqref{prop:htcc:persistence:eq6} holds by definition of the satisfaction relation.
Equivalence between \eqref{prop:htcc:persistence:eq6} and \eqref{prop:htcc:persistence:eq7} holds by definition of $\At{\tuple{h,t}}$.
Equivalence between \eqref{prop:htcc:persistence:eq7} and \eqref{prop:htcc:persistence:eq8} holds by definition of the satisfaction relation.
\end{proof}

\begin{proof}[Proof of Proposition~\ref{prop:arithmetic:persistence}]
 If $h(\eval{h}{t}(\tau))\neq \undefined$, 
 then for all $\cterm{s}{s'}{\varphi}$ occurring in $\tau$,
 either $\tuple{h,t}\models \varphi$ or $\tuple{t,t}\not\models \varphi$ by definition of $\eval{h}{t}$ and application of evaluation $h$ to an arithmetic expression.
 If $\tuple{h,t}\models \varphi$ then $\tuple{t,t}\models \varphi$
 by Proposition~\ref{prop:htcc:persistence}.\ref{item:1:prop:htcc:persistence},
 and therefore $\eval{h}{t}\cterm{s}{s'}{\varphi}=\eval{t}{t}\cterm{s}{s'}{\varphi}$.
 In case that $\tuple{t,t}\not\models \varphi$,
 $\eval{h}{t}\cterm{s}{s'}{\varphi}=\eval{t}{t}\cterm{s}{s'}{\varphi}$ is implied by definition.
 Therefore, $\eval{h}{t}(\tau)=\eval{t}{t}(\tau)$,
 and since $\restr{h}{\vars{\eval{h}{t}(\tau)}}=\restr{t}{\vars{\eval{t}{t}(\tau)}}\neq \undefined$\footnote{Given a valuation~$v$ and set of variables $X \subseteq \X$,
by $\restr{v}{X}$, we denote the restriction of $v$ to $X$, that is, a function
$\restr{v}{X}:X\rightarrow\mathcal{D}_{\undefined}$
such that $v(x) = \restr{v}{X}(x)$ for every variable $x \in X$.} due to $h(\eval{h}{t}(\tau))\neq \undefined$,
 we have $h(\eval{h}{t}(\tau))=t(\eval{t}{t}(\tau))$.
\end{proof}

\begin{proof}[Proof of Proposition~\ref{prop:assign:nondirect}]
 For proving (i), notice that the expression $A \wedge \df{A}$ corresponds to:
\[ 
\neg \neg \df{A} \wedge (\df{A} \rightarrow \alpha \leq {x} \wedge {x} \leq \beta) \wedge \df{A}
\] 
but since $\varphi \models \neg \neg \varphi$ and $\varphi \wedge (\varphi \rightarrow \psi) \equiv \varphi \wedge \psi$ in \HT, the formula above is equivalent to:
\[ 
\alpha \leq {x} \wedge {x} \leq \beta \wedge \df{A}
\]
Finally, as $\alpha \leq {x} \wedge {x} \leq \beta \models \df{A}$ we can remove the conjunct $\df{A}$ above.

For (ii) we have:
\[
\begin{array}{rcl}
\neg A & \equiv & \neg (\neg \neg \df{A} \wedge (\df{A} \rightarrow \alpha \leq {x} \wedge {x} \leq \beta) \\
& \equiv & \neg \neg \neg \df{A} \vee \neg \neg \df{A} \wedge \neg (\alpha \leq {x} \wedge {x} \leq \beta)\\
& \equiv & \neg \df{A} \vee \neg (\alpha \leq {x} \wedge {x} \leq \beta)
\end{array}
\]
But, as $(\alpha \leq {x} \wedge {x} \leq \beta) \models \df A$, we conclude $\neg \df A \models \neg (\alpha \leq {x} \wedge {x} \leq \beta)$ and so the formula above is equivalent to $\neg (\alpha \leq {x} \wedge {x} \leq \beta)$.

\end{proof}

For logic programming syntax, we use comma `$,$' and semicolon `$;$' as alternative representations of $\wedge$ and $\vee$, respectively.
Similarly, we write $\varphi \leftarrow \psi$ to stand for $\psi \rightarrow \varphi$, as expected.
An $\HTC$-literal is either a conditional constraint atom $A$ or its default negation $\neg A$.
An $\HTC$ program is a set of rules of the form:
\begin{align*}
L_1 ; \dots ; L_n \leftarrow L_{n+1},\dots,L_m
\end{align*}
where each $L_i$ is an $\HTC$-literal.
Let $\Head(r)$ stand for the set $\{L_1,\dots,L_n\}$
and $\Body(r)$ be the set $\{L_{n+1},\dots,L_m\}$.
By abuse of notation, we sometimes use~$\Head(r)$
to stand for the disjunction~$\bigvee \Head(r)$
and $\Body(r)$ to stand for the conjunction~$\bigwedge \Body(r)$.

We say that the value of a variable $x \in \X$ is \emph{supported} wrt.\ an $\HTC$-program~$\Pi$ and a valuation~$v$
iff there is a rule~$r \in \Pi$ and a
non-negated constraint atom~$c$ in the head of $r$ satisfying the following conditions:
\begin{enumerate}
\item $x \in \vars{c}$,
\label{def:supported:cond:1}
\item $v \not\models c'$ for every constraint atom~$c'$ in the head of~$r$ such that $x \notin \var(c)$,
\label{def:supported:cond:2}
\item $v \models\Body(r)$.
\label{def:supported:cond:3}
\end{enumerate}
We say that a model~$v$ of some $\HTC$-program~$\Pi$ is \emph{supported}
iff every variable defined is supported wrt.~$\Pi$ and~$v$.

\begin{lemma}\label{lem:supported.aux}
Let $\tuple{t,t}$ be an equilibrium model of some $\HTC$-program~$\Pi$,
$x \in \X$ be some variable which is defined in~$t$
and
$\tuple{h,t}$ be some interpretation with
$h(x) = \undefined$
and
$h(y) = t(y)$ for every variable~$y \in \X \setminus\{x\}$.
Then,
there is a rule~$r \in \Pi$ and a
non-negated constraint atom~$c$ in the head of $r$ satisfying the following conditions:
\begin{enumerate}
\item $x \in \var(c)$,
 \label{lem:supported:cond:1}
\item $\tuple{t,t} \not\models c'$ for every constraint atom~$c'$ in the head of~$r$ such that $x \notin \var(c')$,
 \label{lem:supported:cond:2}
\item $\tuple{h,t} \models \Body(r)$
 \label{lem:supported:cond:3}
\end{enumerate}
\end{lemma}

\begin{proof}[Proof of Lemma~\ref{lem:supported.aux}]
 Let $\tuple{t,t}$ be an equlibrium model of program~$\Pi$,
 $x \in \X$ be some variable which is defined in~$t$
 and
 $\tuple{h,t}$ be some interpretation with
 $h(x) = \undefined$
 and
 $h(y) = t(y)$ for every variable~\mbox{$y \in \X \setminus\{x\}$}.
 We show that $\tuple{h,t}\models\Pi$ whenever one of the conditions \ref{lem:supported:cond:1}-\ref{lem:supported:cond:3} is not fulfilled for $x$,
 thus a contradiction follows with $\tuple{t,t}$ being an equilibrium model,
 and a rule $r$ fulfilling all conditions has to exist.
 \begin{itemize}
   \item[\ref{lem:supported:cond:1}] Assume there exists no $c$ with $x\in\vars{c}$ for $c\in\Head(r)$ and \mbox{$r\in\Pi$}, then
   $\tuple{h,t} \models \Head(r)$ iff $\tuple{t,t} \models \Head(r)$
   for every rule~\mbox{$r \in \Pi$}.
   Furthermore, since $\tuple{t,t}$ is a model of~$\Pi$, it follows that
   $\tuple{t,t} \models \Body(r) \to \Head(r)$ and, thus,
   either
   $\tuple{t,t} \models \Head(r)$
   or
   $\tuple{t,t} \not\models\Body(r)$.
   As we have seen above, the former implies that
   $\tuple{h,t} \models \Head(r)$.
   From Proposition~\ref{prop:htcc:persistence}.\ref{item:1:prop:htcc:persistence},
   the latter implies
   $\tuple{h,t} \not\models\Body(r)$.
   Hence
   $\tuple{h,t} \models \Body(r) \to \Head(r)$.
  This implies that $\tuple{h,t}\models \Pi$ which contradicts the fact that $\tuple{t,t}$ is an equilibrium model of $\Pi$and therefore a rule has to exists with $x$ in the head.
   \item[\ref{lem:supported:cond:2}] Since we proved Lemma~\ref{lem:supported.aux}.\ref{lem:supported:cond:1},
   we only need to consider rules $r$ with $x\in\vars{\Head(r)}$.
   
   Assume there exists a $c\in \Head(r)$ with $x\not\in\vars{c}$ such that $\tuple{t,t}\models c$ for all rules $r\in\Pi$ with $x\in\vars{\Head(r)}$, 
   then we know $\tuple{t,t}\models \Head(r)$ and $\tuple{h,t}\models \Head(r)$,
   due to definition of satisfaction relation for disjunction and Condition~\ref{den:prt:2} for the denotation.
   Thus rule $r$ is fulfilled regardless of the body due to definition of satisfaction relation for implication.
   It follows that $\tuple{h,t}\models\Pi$ which again contradicts $\tuple{t,t}$ being an equilibrium model and therefore there is at least a rule with $x$ in the head
   where no head atoms not containing $x$ is satisfied.
   \item[\ref{lem:supported:cond:3}] 
   We only have to consider rules $r\in\Pi$ with $x\in\vars{\Head(r)}$ and there exists no $c\in \Head(r)$ with $x\not\in\vars{c}$ and $\tuple{t,t}\models c$
   as shown above.

   Assume $\tuple{h,t}\not\models\Body(r)$,
   then rule $r$ is satisfied regardless of the head,
   and therefore $\tuple{h,t}\models\Pi$,
   our final contradiction to $\tuple{t,t}$ being an equilibrium model.\qed
 \end{itemize}
\let\qed\relax
\end{proof}

\begin{proposition}\label{prop:supported}
Every stable model of an $\HTC$ program is also supported.
\end{proposition}

\begin{proof}[Proof of Proposition~\ref{prop:supported}]
 Proposition~\ref{prop:supported} follow directly from Lemma~\ref{lem:supported.aux}, 
 as conditions \ref{def:supported:cond:1} and \ref{def:supported:cond:2} in the definition of supported are identical to conditions \ref{lem:supported:cond:1} and \ref{lem:supported:cond:2} in Lemma~\ref{lem:supported.aux},
 and Condition~\ref{lem:supported:cond:3} and implies $\tuple{t,t}\models \Body(r)$ by Proposition~\ref{prop:htcc:persistence}.\ref{item:1:prop:htcc:persistence}.
 Thus, every stable model $t$ is supported.
\end{proof} 

Note that the proof for Proposition~\ref{prop:lc.supported} follows after the following proofs for Theorem~\ref{th:assign}
as Proposition~\ref{prop:lc.supported} relies on Theorem~\ref{th:assign}.

\begin{lemma}\label{lem:ht:equiv}
The following are valid HT-equivalences:
\begin{align}
\gamma \vee (\varphi \rightarrow \psi) & \equiv (\varphi \rightarrow \psi \vee \gamma) \nonumber\\ & \phantom{\equiv} \wedge (\neg \psi \rightarrow \neg \varphi \vee \gamma) \label{f:orimp}\\
(\varphi \rightarrow (\psi \rightarrow \gamma)) & \equiv (\varphi \wedge \psi \rightarrow \gamma) \label{f:nest-impl}\\
(\varphi \rightarrow \psi \wedge \gamma) & \equiv (\varphi \rightarrow \psi) \wedge (\varphi \rightarrow \gamma) \label{f:andimp}\\
\gamma \vee \neg \neg \varphi & \equiv \neg \varphi \rightarrow \gamma \label{f:negneg} \\
\gamma \vee \neg \neg \varphi \wedge (\varphi \rightarrow \psi) & \equiv (\varphi \rightarrow \psi \vee \gamma) \nonumber\\
& \phantom{\equiv} \wedge (\neg \psi \rightarrow \gamma) \wedge (\neg \varphi \rightarrow \gamma) \label{f:df}\\
\end{align}
\end{lemma}
\begin{proof}[Proof of Lemma~\ref{lem:ht:equiv}]
\eqref{f:orimp} and \eqref{f:nest-impl} are subcases of transformation (R5) in~\cite{capeva05a} whereas \eqref{f:andimp} and \eqref{f:negneg} respectively correspond to (i) and (iv) from Proposition 6 (iv) in~\cite{litatu99a} for nested expressions, which are valid in \HT~\cite{lipeva01a}.
For \eqref{f:df} we apply De Morgan to the left hand side obtaining:
\begin{eqnarray*}
 &  & (\gamma \vee \neg \neg \varphi) \wedge (\gamma \vee (\varphi \rightarrow \psi))\\
 & \equiv & (\gamma \vee \neg \neg \varphi) \wedge (\varphi \rightarrow \gamma \vee \psi) \wedge (\neg \psi \rightarrow \gamma \vee \neg \varphi)
\end{eqnarray*}
where we applied \eqref{f:orimp} in the second conjunct.
By \eqref{f:negneg}, we can replace the first conjunct by $\neg \varphi \rightarrow \gamma$ but then, in the presence of this last conjunct, the consequent of the last implication $\gamma \vee \neg \varphi$ can be replaced by $\gamma$ obtaining:
\begin{eqnarray*}
(\neg \varphi \rightarrow \gamma) \wedge (\varphi \rightarrow \gamma \vee \psi) \wedge (\neg \psi \rightarrow \gamma) \hspace{20pt} \Box
\end{eqnarray*}
\end{proof}

\begin{lemma}\label{lem:oneassig}
Let $A$ be an assignment $\ass{x}{\alpha}{\beta}$. Then $\gamma \vee A$ is equivalent to:
\[
(\df{A} \rightarrow \Phi(A) \vee \gamma) \wedge (\neg \Phi(A) \rightarrow \gamma)
\]
\end{lemma}
 
\begin{proof}[Proof of Lemma~\ref{lem:oneassig}]
By \eqref{f:assig}, $A$ corresponds to the formula $\neg \neg \df{A} \wedge (\df{A} \rightarrow \Phi(A))$.
This formula follows the pattern of the left hand side of \eqref{f:df}, making the replacements $\varphi$ by $\df{A}$ and $\psi$ by $\Phi(A)$.
As a result, we obtain that $A$ is equivalent to:
\begin{eqnarray*}
(\df{A} \rightarrow \Phi(A) \vee \gamma) \wedge (\neg \df{A} \rightarrow \gamma) \wedge (\neg \Phi(A) \rightarrow \gamma)
\end{eqnarray*}
But now, we observe that $\Phi(A) \models \df{A}$ since satisfying $\alpha \leq x \wedge x \leq \beta$ always implies satisfying $\df{\alpha}$ and $\df{\beta}$.
Since HT satisfies contraposition, $\neg \df{A} \models \neg \Phi(A)$ and so $(\neg \df{A} \rightarrow \gamma)$ is subsumed by $(\neg \Phi \rightarrow \gamma)$, so that we can remove the former.
\end{proof}

\begin{proof}[Proof of Theorem~\ref{th:assign}]
For any $i=0,\dots,n$ let $\Head_i$ to stand for the set $\{A_1,\dots,A_i\}$.
Note that when $i=0$, $\Head_i=\emptyset$.
We prove that by induction $i$ that \eqref{f:rule} is equivalent to the set of rules $S_i$ defined as:
\begin{multline}
\gamma_i \vee \bigvee_{A \in \Delta} \Phi(A) \leftarrow \\
\bigwedge_{A\in\Body(r)} 
\!\!A
 \wedge \bigwedge_{A \in \Delta} \df{A} \wedge \bigwedge_{A' \in \Head_i \setminus \Delta} \neg \Phi(A') \label{f:hdi}
\end{multline}
for all $\Delta \subseteq \Head_i$, where $\gamma_i$ stands for $A_{i+1}\vee \dots \vee A_n$.
For $i=0$ we have that $\gamma_i=Head(r)$ and $\Head_i=\emptyset$ so its unique subset is $\Delta=\emptyset$ and the expression above trivially amounts to \eqref{f:rule} (empty disjunctions and conjunctions respectively amount to $\bot$ and $\top$, as usual).
For the inductive step, assume it holds for $0\leq i<n$ and we want to prove it for $i+1$.
Take any rule like \eqref{f:hdi} in $S_i$ for some fixed $\Delta \subseteq \Head_i$. Since $i<n$, $\gamma_i = A_{i+1} \vee \gamma_{i+1}$. If we apply Lemma~\ref{lem:oneassig} on the head of \eqref{f:hdi} taking $\gamma = \gamma_{i+1} \vee \bigvee_{A \in \Delta} \Phi(A)$ and $A = A_{i+1}$ we obtain the conjunction of the two implications:
\begin{eqnarray*}
\Phi(A_i) \vee \gamma_{i+1} \vee \bigvee_{A \in \Delta} \Phi(A) & \leftarrow & \df{A_i}\\
\gamma_{i+1} \vee \bigvee_{A \in \Delta} \Phi(A) & \leftarrow & \neg A_i
\end{eqnarray*}
in the head of the rule.
Now, using \eqref{f:andimp} to split the conjunction in the head into two different implications, and \eqref{f:nest-impl} to remove nested implications, we get the pair of rules:
\begin{multline}
\gamma_{i+1} \vee \bigvee_{A \in \Delta \cup \{A_i\}} \Phi(A) \leftarrow \\ \bigwedge_{A\in\Body(r)} 
\!\!A
 \wedge \bigwedge_{A \in \Delta \cup \{A_i\}} \df{A} \wedge \bigwedge_{A' \in \Head_i \setminus \Delta} \neg \Phi(A') \label{f:hdi1a}
\end{multline}
\begin{multline}
\gamma_{i+1} \vee \bigvee_{A \in \Delta} \Phi(A) \leftarrow \\ \bigwedge_{A\in\Body(r)} 
\!\!A
 \wedge \bigwedge_{A \in \Delta} \df{A} \wedge \bigwedge_{A' \in \{A_i\}\cup \Head_i \setminus \Delta} \neg \Phi(A') \label{f:hdi1b}
\end{multline}
It is not difficult to see that these two rules belong to $S_{i+1}$ and respectively correspond to the subsets $\Delta \cup \{A_i\}$ and $\Delta$ of $\Head_{i+1}$ -- notice that 
 $\Head_{i+1} \setminus (\Delta \cup \{A_i\}) = \Head_i \setminus \Delta$.
Moreover, for any rule in $S_{i+1}$ fixing some $\Delta' \subseteq \Head_{i+1}$, we may find the corresponding rule in $S_i$ with $\Delta=\Delta' \setminus \{A_i\}$ so that splitting the latter generates the former.
Therefore, using this splitting for each rule in $S_i$ we get exactly all rules in $S_{i+1}$, and the inductive step is proved.

Finally, it simply remains to observe that the set of rules in the enunciate of the Theorem corresponds to the case $i=n$, where $\gamma_i = \top$ (the empty disjunction) and $\Head_i = \Head(r)$.

\end{proof}

\begin{lemma}\label{lem:lc.supported.aux}
Let and $\tuple{t,t}$ be an equilibrium model of some~$\Pi$ be an~$\LC$-program,
$x \in \X$ be some variable with $t(x)=d\in\mathcal{D}$
and
$\tuple{h,t}$ be some interpretation with
$h(x) = \undefined$
and
$h(y) = t(y)$ for every variable~$y \in \X \setminus\{x\}$.
Then, 
there is a rule~$r \in \Pi$ and an assignment $A$ of the form~$x := \alpha..\beta$
in the head of $r$ satisfying the following conditions:
\begin{enumerate}
\item $v(x)=d$ and $v(\alpha)\leq d \leq  v(\beta)$, so both $v(\alpha), v(\beta) \in \mathbb{Z}$
\item $\tuple{t,t} \not\models A'$ for every assignment of the form
$y:=\alpha'..\beta'$ in the head of~$r$ with $x\neq y$,

\item $\tuple{h,t} \models \Body(r)$.
\end{enumerate}
\end{lemma}

\begin{proof}[Proof of Lemma~\ref{lem:lc.supported.aux}]
Let~$\Pi$ be any~$\LC$-program,
$\tuple{t,t}$ be some equilibrium model of~$\Pi$
and $\Pi'$ be its corresponding~$\HTC$-program obtained as outlined in Theorem~\ref{th:assign}.
Then, $\tuple{t,t}$ is also an equilibrium model of~$\Pi'$.
Let $\tuple{h,t}$ be some interpretation with
$h(x) = \undefined$
and
$h(y) = t(y)$ for every variable~$y \in \X \setminus\{x\}$.
From Lemma~\ref{lem:supported.aux},
there is a rule~$r' \in \Pi'$ and a
non-negated constraint atom~$c$ in the head of $r'$ satisfying the following conditions:
\begin{enumerate}
\item $x \in \vars{c}$,

\item $\tuple{t,t} \not\models c'$ for every constraint atom~$c'$ in the head of~$r'$ such that $x \notin \vars{c}$,

\item $\tuple{h,t} \models \Body(r')$.
\end{enumerate}
By construction, this rule~$r'$ must correspond to some rule~$r \in \Pi$ of the form
\begin{gather}
x_1 := \alpha_1..\beta_1
\vee \dotsc \vee
x_n := \alpha_n..\beta_n
\leftarrow B
\end{gather}
and $c$ must either be of the form $\alpha_i \leq x_i$ or $x_i \leq \beta_i$.
Note that the body of $r$ is a conjunction of literal that contains all literals in the body of $r'$
and, thus, it immediately follows that
\mbox{$\tuple{h,t} \models \Body(r)$}.

Now, we show that Condition~2 holds.
Suppose, that this is not the case.
Then we can show that $\tuple{h,t}\models \Pi$, which is a contradiction to $t$ being a stable model.
Note that we only have to examine rules $r\in \Pi$
where there exists an assignment $A'=x':=\alpha'..\beta'\in \Head(r)$ with $x\ne x'$, $x\in\vars{\alpha',\beta'}$ and $t\models x':=\alpha'..\beta'$,
and $\tuple{h,t}\models\Body(r)$.
If the former does not hold, 
then by assumption there exists an Assignment $x'':=\alpha''..\beta''\in \Head(r)$ with $x\not\in\vars{\alpha',\beta'}$ and $t\models x'':=\alpha''..\beta''$,
therefore $\tuple{h,t}\models x'':=\alpha''..\beta''$ as $x$ has no impact on satisfaction, and finally $\tuple{h,t}\models r$.
If the ladder is not the case, $\tuple{h,t}\models r$ due to $\tuple{h,t}\not\models\Body(r)$.
Then, if $x\in \vars{\eval{h}{t}(\alpha'),\eval{h}{t}(\beta')}$,
$\tuple{h,t}\not\models \df{A'}$ but $\tuple{h,t}\models\neg\neg \df{A'}$ due to $t\models\df{A'}$.
Therefore, by definition of assignment $\tuple{h,t}\models A'$.
If $x\not\in \vars{\eval{h}{t}(\alpha'),\eval{h}{t}(\beta')}$,
then either $\eval{h}{t}(\alpha')=\eval{t}{t}(\alpha')$ or $\eval{h}{t}(\alpha')=\undefined$,
and $\eval{h}{t}(\beta')=\eval{t}{t}(\beta')$ or $\eval{h}{t}(\beta')=\undefined$, respectively.
If both are equally, naturally $\tuple{h,t}\models A'$,
and if either is undefined it holds again that $\tuple{h,t}\not\models \df{A'}$ but $\tuple{h,t}\models\neg\neg \df{A'}$,
and therefore $\tuple{h,t}\models A'$.
In each case, it therefore follows that $\tuple{h,t}\models \Head(r)$ and thus $\tuple{h,t}\models r$.
Now we have that $\tuple{h,t}\models\Pi$ and thus a contradiction to $t$ being a stable model of $\Pi$.
 
Finally, we show Condition~1.
We have $t\models \Body(r)$ by $\tuple{h,t}\models \Body(r)$ and Proposition~\ref{prop:htcc:persistence}.\ref{item:1:prop:htcc:persistence},
therefore, $t\models \Head(r)$ since $t\models r$,
and due to Condition~2,
we know there exist an $A=x:= \alpha..\beta\in\Head(r)$, 
such that $t\models x := \alpha..\beta$ as no assignment of a variable $y\ne x$ can satisfy the head.
Finally, $t\models x := \alpha..\beta$ implies
$t\models \df{\alpha}\wedge\df{\beta}$ and $t\models \alpha\leq x\wedge x\leq\beta$,
and therefore, $t(x)=d\in\mathcal{D}$ with $t(\alpha)\leq d \leq t(\beta)$.
Hence, the result holds.
\end{proof}
\begin{proof}[Proof of Proposition~\ref{prop:lc.supported}]
 Proposition~\ref{prop:lc.supported} follow directly from Lemma~\ref{lem:lc.supported.aux}, 
 as conditions 1 and 2 in the definition of supported are identical to conditions 1 and 2 in Lemma~\ref{lem:supported.aux},
 and Condition~\ref{lem:supported:cond:3} implies $\tuple{t,t}\models \Body(r)$ by Proposition~\ref{prop:htcc:persistence}.\ref{item:1:prop:htcc:persistence}.
 Thus, every stable model $t$ is supported.
\end{proof}

\begin{lemma}\label{lem:gz.translation.aux2}
Let $\Gamma$ be some theory,
$\tau$ be some conditional expression
and $\tuple{h,t}$ be some model of~$\Gamma$.
Then,
$\tuple{h,t}_\tau$ is a model of~$\Gamma \cup \{ \delta(\tau) \}$.
\end{lemma}

\begin{proof}[Proof of Lemma~\ref{lem:gz.translation.aux2}]
We proceed by cases.
\begin{enumerate}
\item We assume first that $\tuple{h,t} \models \varphi$.
Then, it immediately follows that $\tuple{h,t}_\tau$ satisfies
\eqref{eq:gz.translation.5}-\eqref{eq:gz.translation.4}.
Furthermore, by construction, we can see that
$v_\tau(x_\tau) = v(s_1)$ with $v \in \{ h, t \}$.
Therefore, $\tuple{h,t}_\tau$ also satisfies
\eqref{eq:gz.translation.1}-\eqref{eq:gz.translation.2}.

\item In case that $\tuple{h,t} \models \neg\varphi$,
we immediately get that
 $\tuple{h,t}_\tau$ satisfies
\eqref{eq:gz.translation.1}-\eqref{eq:gz.translation.2}
and
\eqref{eq:gz.translation.5}.
Furthermore, by construction, we can see that
$v_\tau(x_\tau) = v(s_2)$ with $v \in \{ h, t \}$
and, thus, $\tuple{h,t}_\tau$ also satisfies
\eqref{eq:gz.translation.3}-\eqref{eq:gz.translation.4}.

\item Finally, in case that $\tuple{h,t} \not\models \varphi$
and $\tuple{h,t} \not\models \neg\varphi$,
we get that $\tuple{t,t} \models \varphi$.
Hence,
$\tuple{h,t}_\tau$ satisfies
\eqref{eq:gz.translation.3}-\eqref{eq:gz.translation.4}.
Furthermore, by construction, we get that 
$t_\tau(x_\tau) = t(s_1) = t_\tau(s_1)$
and, since $\tuple{h,t} \not\models \varphi$,
this implies that
\eqref{eq:gz.translation.1}-\eqref{eq:gz.translation.2}.
Also by construction, we get $h_\tau(x_\tau) = \undefined$
which, together with $\tuple{t,t} \models\varphi$, implies that
$\tuple{h,t}_\tau$ satisfies~\eqref{eq:gz.translation.5}.\qed
\end{enumerate}
\let\qed\relax
\end{proof} 

\begin{proof}[Proof of Proposition~\ref{prop:translation.model.characterisation}]
We proceed by cases.
\begin{enumerate}
\item We assume first that $\tuple{h,t} \models \varphi$.
Then, since $\tuple{h,t}$ is a model of~$\delta(\tau)$,
it follows that it satisfies
$x_\tau = s_1 \leftarrow \varphi \wedge \df{s_1}$.
Hence, either
$h(x_\tau) = h(s_1) \neq \undefined$
or
$\tuple{t,t} \not\models \df{s_1}$
or
both
$t(x_\tau) = t(s_1) \neq \undefined$
and
$\tuple{h,t} \not\models \df{s_1}$.
In the first case the result holds.
Otherwise,
$\tuple{v,t} \not\models \df{s_1}$ with $v = t$ (resp. $v = h$).
This implies that $v(s_1) = \undefined$
and, since $\tuple{h,t}$ satisfies
$x_\tau = s_1 \leftarrow \varphi \wedge \df{x_\tau}$,
we get that $\tuple{v,t} \not\models \df{x_\tau}$.
That is, $v(x_\tau) = \undefined = v(s_1)$.
Hence, if $v = t$, we get that
$t(x_\tau) = t(s_1) = \undefined$ which, in its turn implies
$h(x_\tau) = h(s_1) = \undefined$
and the result holds.
Otherwise, $v = h$,
we only get
$h(x_\tau) = h(s_1) = \undefined$.
However, $v = h$ implies that we are in the third case and then
$t(x_\tau) = t(s_1) \neq \undefined$ and the result also hold.

\item The case in which $\tuple{h,t} \models \neg\varphi$
is analogous.

\item Finally, assume that
\mbox{$\tuple{h,t} \not\models \varphi$}
and
\mbox{$\tuple{h,t} \not\models \neg\varphi$}.
Then, it follows that
\mbox{$\tuple{t,t} \models \varphi$}
and, from the first point above, we get that $t(x_\tau) = t_\tau(x_\tau) = t(s_1)$.
Furthermore, since $\tuple{h,t}$ satisfies
$\varphi \leftarrow \neg\neg\varphi \wedge \df{x_\tau}$
this implies that
$h(x_\tau) = \undefined$
and the lemma holds.\qed
\end{enumerate}
\let\qed\relax
\end{proof}
\begin{proof}[Proof of Proposition~\ref{prop:translation.theory.equiv.proj}]
Let~$\Delta$ be any set of formulas over~$X$
and assume that $t$ is a stable model of~$\Gamma \cup \Delta$.
Then, from Lemma~\ref{lem:gz.translation.aux2},
we get that $\tuple{t,t}_\tau$ is a model of $\Gamma \cup \Delta \cup \{ \delta(\tau) \}$
and, by construction, we can see that $\restr{t}{X} = \restr{t_\tau}{X}$.
Suppose that this is not an equilibrium model, that is,
that there is some model $\tuple{h',t_\tau}$ of $\Gamma \cup \Delta \cup \{ \delta(\tau) \}$
such that $h' \subset t_\tau$.
Let $h$ be a valuation such that
$\restr{h}{X} = \restr{h'\!}{X}$
and
$h(x_\tau) = \undefined$.
From Lemma~\ref{prop:translation.model.characterisation},
we get that $h'=h'_\tau$
and, thus, we can see that $h'=h'_\tau = h_\tau$.
Note that $h' \subset t_\tau$ and $\restr{t_\tau}{X}=\restr{t}{X}$
imply that $h \subseteq t$.
Furthermore, since $t$ is a stable model,
it must be that $h = t$.
This implies that $h_\tau = t_\tau$
and, since $h' = h_\tau$, that $h' = t_\tau$.
This is a contradiction with the fact that
$h' \subset t_\tau$.
Consequently, 
$\tuple{t,t}_\tau$ is a model of $\Gamma \cup \Delta \cup \{ \delta(\tau) \}$.
\\[5pt]
The other way around.
Let $t$ be a stable model of~$\Gamma \cup \Delta \cup \{ \delta(\tau) \}$
and let $t'$ be a valuation such that
$\restr{t'\!}{X} = \restr{t}{X}$
and
$t'(x_\tau) = \undefined$.
Then,
since $x_\tau$ does not occur in~$\Gamma \cup \Delta$,
it follows that~$\tuple{t',t'}$ is a total model of $\Gamma \cup \Delta$.
Suppose that this is not an equilibrium model, that is,
that there is some model $\tuple{h,t'}$ of $\Gamma \cup \Delta$
such that $h \subset t'$.
From Lemma~\ref{prop:translation.model.characterisation},
it follows that $t = t_\tau = t'_\tau$
and, from Lemma~\ref{lem:gz.translation.aux2},
we get that $\tuple{h,t}_\tau \models \Gamma \cup \Delta \cup \{ \delta(\tau) \}$.
Note that $\tuple{h,t}_\tau = \tuple{h_\tau,t_\tau} = \tuple{h_\tau,t}$
and, thus,
$\tuple{h_\tau,t} \models \Gamma \cup \Delta \cup \{ \delta(\tau) \}$.
Furthermore, since  $t$ is a stable model of~$\Gamma \cup \Delta \cup \{ \delta(\tau) \}$,
it follows that $h_\tau = t$.
Hence,
we get
$\restr{h}{X} = \restr{h_\tau}{X} = \restr{t}{X} = \restr{t'\!}{X}$.
However, this is a contradiction with the fact that 
$h \subset t'$
and
$t'(x_\tau) = \undefined$.
\end{proof}

\begin{proof}[Proof of Theorem~\ref{thm:translation.theory.equiv}]
Combining Corollary~\ref{cor:translation.theory.equiv}
and Proposition~\ref{prop:translation.theory.equiv.proj}
we immediately get
$\Gamma \equiv_s^X \Gamma \cup \{ \delta(\tau) \} \equiv \Gamma[\tau/x_\tau] \cup \{ \delta(\tau) \}$.
\end{proof} 


\end{document}